\documentclass{article}

\PassOptionsToPackage{numbers, compress}{natbib}

\usepackage[preprint]{neurips_2025}
\usepackage[utf8]{inputenc} % allow utf-8 input
\usepackage[T1]{fontenc}    % use 8-bit T1 fonts
\usepackage{hyperref}       % hyperlinks
\usepackage{url}            % simple URL typesetting
\usepackage{booktabs}       % professional-quality tables
\usepackage{amsfonts}       % blackboard math symbols
\usepackage{nicefrac}       % compact symbols for 1/2, etc.
\usepackage{microtype}      % microtypography
\usepackage[table]{xcolor}         % colors

\usepackage{tikz}
\usetikzlibrary{shapes,arrows,positioning,fit,backgrounds}
\usepackage{graphicx}
\graphicspath{ {./} }

\usepackage[acronym]{glossaries}
\makeglossaries

\usepackage{float}
\usepackage{amsmath}
\usepackage{caption}
\usepackage{subcaption}
\usepackage{tcolorbox}

\newacronym{map}{mAP}{mean Average Precision}
\newacronym{ap}{AP}{Average Precision}
\newacronym{cot}{CoT}{Chain-of-Thought}
\newacronym{llm}{LLM}{Large Language Model}
\newacronym{lora}{LoRA}{Low-Rank Adaptation}
\newacronym{lvlm}{LVLM}{Large Vision-Language Model}
\newacronym{rnn}{RNN}{Recurrent Neural Network}

\title{StoryReasoning Dataset: Using Chain-of-Thought for Scene Understanding and Grounded Story Generation}

\author{%
    Daniel A. P. Oliveira \\
    Instituto Superior Técnico, Universidade de Lisboa\\
    INESC-ID Lisboa\\
    \texttt{daniel.oliveira@inesc-id.pt} \\
    \And
    David Martins de Matos \\
    Instituto Superior Técnico, Universidade de Lisboa\\
    INESC-ID Lisboa\\
    \texttt{david.matos@inesc-id.pt} \\
}

\begin{document}
    \maketitle

    \begin{abstract}
        Visual storytelling systems struggle to maintain character identity across frames and link actions to appropriate subjects,
        frequently leading to referential hallucinations.
        These issues can be addressed through grounding of characters, objects, and other entities on the visual elements.
        We propose StoryReasoning, a dataset containing 4,178 stories derived from 52,016 movie images,
        with both structured scene analyses and grounded stories.
        Each story maintains character and object consistency across frames while explicitly modeling multi-frame relationships
        through structured tabular representations.
        Our approach features cross-frame object re-identification using visual similarity and face recognition,
        chain-of-thought reasoning for explicit narrative modeling, and a grounding scheme that links textual
        elements to visual entities across multiple frames.
        We establish baseline performance by fine-tuning Qwen2.5-VL 7B, creating Qwen Storyteller, which performs end-to-end object detection,
        re-identification, and landmark detection while maintaining consistent object references throughout the story.
        Evaluation demonstrates a reduction from 4.06 to 3.56 (-12.3\%) hallucinations on average per story and an improvement
        in creativity from 2.58 to 3.38 (+31.0\%) when compared to a non-fine-tuned model.
    \end{abstract}

    \section{Introduction}\label{sec:introduction}
    Visual storytelling the generation of narratives from image sequences, represents a challenge at the intersection of computer vision
    and natural language processing with applications in entertainment, education, and journalism~\cite{Oliveira2024StoryGF}.
    While significant progress has been made in image captioning~\cite{li2023blip, zhang2024groundhog} and object
    detection~\cite{qwen25vl, cheng2021mask2former}, current approaches face challenges when generating coherent stories
    from image sequences~\cite{Oliveira2024StoryGF}.
    These challenges include maintaining consistent object identity, grounding story elements to visual entities, and constructing coherent narratives.
    Existing visual storytelling systems often produce descriptions that suffer from inconsistent character references, contradictory object relationships,
    and a lack of coherent narrative structure~\cite{tacl_a_00553}.
    Even state-of-the-art \glspl{lvlm} struggle with consistent character references, link actions to the appropriate subjects, and frequently suffer from hallucinations~\cite{farquhar2024, Liu2024ASO}.
    To address these limitations in visual storytelling, we build upon our previous work on GroundCap~\cite{Oliveira2025GroundCapAV},
    which introduced ID-based grounding for single images.
    While GroundCap demonstrated the effectiveness of structured entity references within individual frames, the challenge of
    maintaining consistent identity across multiple images requires extending this approach to sequential contexts.
    This natural progression led us to develop the Story Reasoning dataset.
    This dataset specifically designed for multi-frame narrative
    generation that preserves the strengths of our single-image grounding while adding the temporal dimensions.
    Story Reasoning contains 4,178 cohesive stories derived from 52,016 images in the GroundCap dataset, organizing temporally
    connected image sequences extracted from the same movie scenes to ensure narrative coherence.
    The dataset maintains character and object consistency across frames
    while explicitly modeling multi-frame relationships through structured tabular representations of characters, objects, settings, and narrative phases.

    Based on the Story Reasoning dataset, we develop Qwen Storyteller, a model that leverages \gls{cot} reasoning~\cite{wei2022chain, zhang2023multimodal}
    to generate coherent visual narratives.
    The core of our approach is a structured reasoning process for explicit cross-frame object re-identification.
    While current approaches~\cite{qwen25vl} can detect objects within individual images, Qwen Storyteller extends this capability
    by re-identifying these objects across multiple images in a sequence without requiring separate object detectors.
    This enables our system to recognize when a character, person, or object in one frame is the same entity appearing in subsequent frames,
    maintaining consistent identity throughout the narrative.
    For a set of input images, our model produces two key outputs:
    (1) a detailed structured analysis that systematically documents characters, objects, settings, and narrative progression across frames,
    drawing inspiration from graph-based approaches for event ratiocination~\cite{pmlr-v139-zheng21b}; and
    (2) a grounded narrative that uses specialized tags to link story elements directly to visual entities.
    By explicitly modeling the reasoning process, Qwen Storyteller maintains consistent character identities and relationships throughout the narrative,
    ensuring that pronouns and references properly correspond to their visual counterparts and reducing ambiguity and hallucination.
    Our dataset, Story Reasoning, establishes ground truth object identities across frames, while our trained model performs these tasks end-to-end at inference time.
    We evaluate our approach through both automated metrics and using \glspl{llm} for assessment of generated stories,
    providing insights that complement traditional metrics.

    The main contributions of our work include:
    (1) the Story Reasoning dataset comprising 4,178 stories with both structured scene analyzes and grounded stories, and this is, to the best of our knowledge, the first of its kind;
    (2) a \gls{cot} framework for scene understanding that explicitly models cross-frame consistency, character relationships, object interactions, scene settings, and narrative structure through structured tabular representations;
    (3) the Qwen Storyteller model that performs end-to-end object detection, re-identification, and landmark detection while maintaining consistent object references throughout generated narratives;
    (4) an interactive visualization interface that enables real-time story generation and interpretation, with color-coded entity tags and direct visual grounding that makes the connections between narrative elements and visual entities explicit and accessible to users.

    The remainder of this paper is organized as follows: Section~\ref{sec:related-work} reviews related work,
    Section~\ref{sec:the-story-reasoning-dataset} describes the Story Reasoning dataset,
    Section~\ref{sec:qwen-storyteller-the-baseline-model} details our proposed model named Qwen Storyteller,
    Section~\ref{sec:evaluation-results} presents evaluation results,
    Section~\ref{sec:limitations} discusses limitations,
    and Section~\ref{sec:conclusions} provides conclusions and outlines future work.

    \section{Related Work}\label{sec:related-work}
    This section overviews relevant advances in image/video captioning, visual storytelling, chain-of-thought reasoning, and cross-frame consistency techniques.

    \subsection{Image and Video Captioning and Large Vision Language Models}\label{subsec:image-video-captioning-lvlms}
    Visual story generation extends beyond captioning by constructing coherent narratives across images and
    differs from video captioning by adding creativity through speculation about character emotions, motivations, and events while maintaining visual fidelity.

    Recent approaches in image captioning have made substantial progress in grounding textual descriptions to specific visual regions.
    BLIP-2~\cite{li2023blip} leverages frozen image encoders and large language models for improved visual-language alignment,
    while GROUNDHOG~\cite{zhang2024groundhog} introduces pixel-level grounding of text spans to image regions.
    GroundCap~\cite{Oliveira2025GroundCapAV} further extends these capabilities by introducing an ID-based grounding system that enables
    consistent object reference tracking and action-object linking within individual images.

    In video captioning, approaches such as VATEX~\cite{wu2023vatex} address the challenge of describing actions and events over time.
    However, these methods typically focus on generating isolated descriptions of directly observable content in short video clips rather than
    creating cohesive narratives that maintain character and object consistency across longer sequences.
    Unlike video captioning, story generation speculates on characters' emotions, motivations, and underlying narrative events while remaining
    grounded in the visual content.
    This balance allows for creative interpretation {--} such as inferring a character's thoughts or past experiences {--} while ensuring fidelity to
    visible elements (e.g., not depicting a character as happy when they are clearly crying in the frame)~\cite{Oliveira2024StoryGF}.

    Recent \glspl{lvlm} like GPT-4o~\cite{openai2024gpt4o}, Claude 3.7 Sonnet~\cite{anthropic2025}, and Gemini Flash 2.5~\cite{google2025gemini}
    demonstrate strong visual understanding capabilities but they lack the ability to generate precise
    bounding box coordinates for objects they reference in images.
    This limitation impacts their ability to maintain consistent object references, particularly across multiple frames.
    Qwen2.5-VL~\cite{qwen25vl} introduced object grounding capabilities, enabling detection of objects within images
    allowing it to perform visual grounding that other general-purpose models cannot in single images.

    \subsection{Visual Storytelling}\label{subsec:visual-storytelling}
    Visual storytelling extends beyond captioning by generating narratives that connect multiple images through causal and temporal relationships.
    Early approaches such as~\cite{huang2016visual} used sequential \gls{rnn} architectures to generate stories from image sequences but struggled
    with maintaining character consistency and producing coherent narratives.

    Subsequent work has focused on improving narrative coherence through hierarchical approaches and attention mechanisms.
    TAPM~\cite{Yu_2021_CVPR} introduced a transitional adaptation method to better align visual and textual information in generated stories.
    Character-focused approaches such as CharGrid~\cite{tacl_a_00553} implicitly model characters and their relationships across frames,
    while StoryDiffusion~\cite{yang2023storydiffusion} employs consistent self-attention mechanisms to maintain visual coherence in storyboard generation.
    While these methods improve narrative quality or perform implicit character modeling,
    they do not explicitly track object identities across frames or model the narrative in a structured manner.

    \subsection{Chain-of-Thought Reasoning}\label{subsec:chain-of-thought-reasoning}
    \glsreset{cot}
    \gls{cot} reasoning has emerged as a powerful approach for complex problem-solving with \glspl{llm}~\cite{wei2022chain}.
    By breaking down complex tasks into interpretable steps, \gls{cot} methods improve model reasoning and provide explicit rationales
    for generated outputs.
    Multimodal \gls{cot} reasoning~\cite{zhang2023multimodal} extends this approach to visual inputs,
    enabling models to perform structured reasoning about visual content.

    In visual storytelling, graph-based approaches like HEGR~\cite{pmlr-v139-zheng21b} use hypergraphs to model relationships
    between scene elements within single images.
    These methods do not maintain explicit object identity across frames or provide fine-grained grounding of narrative elements to visual regions.

    \subsection{Cross-Frame Consistency in Visual Narratives}\label{subsec:cross-frame-consistency}
    Maintaining consistency across frames remains a significant challenge in visual storytelling.
    Recent work has explored various approaches to address this issue, with particular focus on character-centric modeling and creative story generation.
    Character consistency has been a focus of recent research. Chen et al.~\cite{chen2022character} propose a two-stage
    framework for story visualization (text-to-image generation) that maintains character consistency across generated frames.
    While their approach operates in the reverse direction of our work (generating images from text rather than text from images),
    it highlights the importance of explicit character modeling for narrative coherence in sequential visual content.
    Park et al.~\cite{park2023character} introduce a Character-centric Creative story generation via Imagination (CCI)
    framework that enhances narrative depth through visual representations of key story elements, demonstrating how multimodal
    approaches can improve character development and consistency in generated stories.
    Explicitly modeling character and object identity across frames has been shown to improve narrative coherence and reduce
    hallucinations~\cite{reflect2023, farquhar2024}. REFLECT~\cite{reflect2023} introduces a framework that leverages \glspl{llm}
    for reasoning based on hierarchical summaries of past experiences, while Farquhar et al.~\cite{farquhar2024} propose entropy-based
    uncertainty estimators to detect hallucinations in generated content.
    However, most existing approaches either focus on visual consistency without explicit language grounding or generate
    grounded language without maintaining consistent visual identities across frames.
    Our work bridges this gap by introducing a framework that explicitly models cross-frame object consistency
    while grounding generated narratives to specific visual elements.
    By combining structured chain-of-thought reasoning with consistent object identity tracking, our approach enables the
    generation of coherent stories that maintain character and object consistency while providing explicit visual grounding.

    \section{The Story Reasoning Dataset}\label{sec:the-story-reasoning-dataset}
    The Story Reasoning dataset was created to address limitations in visual storytelling by providing a comprehensive framework for
    maintaining consistent identity across multiple images while generating coherent narratives.
    Story Reasoning contains 4,178 cohesive stories derived from 52,016 images in the GroundCap dataset,
    organizing temporally connected image sequences extracted from the same movie scenes to ensure narrative coherence.
    Each story maintains character and object consistency across frames while explicitly modeling multi-frame relationships
    through structured tabular representations.
    This section details the dataset construction process, from frame selection to \gls{cot} and story generation.

    \subsection{Frame Selection and Organization}
    To create coherent sequences, we select at least 5 images from the same movie and sequential scenes, empirically determined
    to be sufficient for meaningful story creation while preserving narrative continuity.
    The GroundCap~\cite{Oliveira2025GroundCapAV} dataset itself was created using a systematic frame extraction process
    that selected character-containing scenes and extracted middle frames from each shot, yielding 52,016 frames after quality filtering.
    We enforce a character presence requirement—each story must include at least one frame with a person detected to maintain
    the character-centric nature of the narratives.
    This methodology produced 4,178 distinct sets of frames, each containing temporally related images with consistent character
    presence.

    \subsection{Object Detection}\label{subsec:object-detection}
    Our dataset leverages the existing object detections from the GroundCap dataset, which utilizes Mask2Former~\cite{cheng2021mask2former}
    with a Swin-Large backbone~\cite{liu2021swin}.
    This approach enables simultaneous detection and segmentation of both distinct objects (``thing'' classes) and background elements (``stuff'' classes).
    The detection process in GroundCap handles ``thing'' and ``stuff'' classes differently.
    For distinct objects, bounding boxes are directly computed using a greedy approach.
    For background elements like sky or buildings, a K-means clustering approach generates multiple bounding boxes, allowing more granular reference to background elements.

    \subsection{Landmark Detection}\label{subsec:landmark-detection}
    After the object detection step, to enhance setting, location and cultural context, we incorporated a landmark detection using a dual-approach system.
    The primary method utilizes a fine-tuned Swin Transformer~\cite{liu2021swin} model trained on the Google Landmarks Dataset v2~\cite{weyand2020google},
    which contains over 5M images spanning 200k distinct landmarks including architectural structures, natural landmarks, and notable locations.
    This approach provides high-precision landmark recognition with standardized naming conventions.

    When the transformer-based detector fails to identify landmarks (confidence below a threshold of 0.5),
    the system falls back to an \gls{llm}-based detector that uses Qwen-VL 2.5~\cite{qwen25vl} to identify potential landmarks through structured prompting.
    The system provides candidate bounding boxes from previously detected ``stuff'' classes
    (background elements like buildings, walls, or structures) to the \gls{llm}.
    The \gls{llm} then analyzes the image and matches identified landmarks with the most appropriate existing bounding box,
    ensuring that landmarks have accurate spatial information without requiring redundant detection.
    This approach efficiently repurposes generic ``building'' or ``structure'' detections into specific named landmarks
    (e.g., ``Eiffel Tower'' or ``Golden Gate Bridge'').
    This complementary approach enables landmark identification even for less common or region-specific locations that might not be
    well-represented in the training dataset.

    \subsection{Cross-Frame Object Re-identification}\label{subsec:cross-frame-matching}
    For consistent character and object identity across frames, our re-identification system combines visual embeddings
    with face recognition to track entities throughout image sequences.
    Existing tracking systems based on Kalman filters~\cite{welch1995introduction} are unsuitable for our non-continuous movie frames
    spanning different settings, lighting conditions, and viewpoints.
    With scene cuts and changing perspectives, the same object can appear from drastically different angles in consecutive images,
    making motion-prediction approaches ineffective.

    Our approach focuses on visual similarity rather than motion prediction, combining general object embeddings with specialized face embeddings for person detections.
    We evaluated several vision embedding models including OpenAI CLIP~\cite{radford2021learning}, LAION CLIP~\cite{schuhmann2022laion},
    DINOv2~\cite{oquab2023dinov2}, and SigLIP~\cite{zhai2023sigmoid}.
    Empirical testing showed SigLIP with SoViT-400M~\cite{alabdulmohsin2023getting} backbone and 384-pixel patches provides
    superior cross-frame matching despite having fewer parameters than alternatives.
    For each detected object, we compute visual embeddings by first applying Mask2Former~\cite{cheng2021mask2former}
    segmentation to isolate the object from background elements, cropping the precisely segmented region,
    then processing this masked region with SigLIP to generate a normalized embedding vector.
    For persons, we identify face regions and apply dual filtering based on confidence threshold and minimum resolution (128 pixels).
    ArcFace~\cite{arcface2018} embeddings are extracted only from faces exceeding this threshold to avoid identity confusion from unreliable small-face embeddings.

    The matching algorithm processes detections grouped by object class across all frames, meaning that objects of different classes can not be matched.
    Face similarity is prioritized for person detections while falling back to visual similarity when faces are not visible.
    The system employs adaptive thresholding that considers both absolute similarity values and statistical distribution of
    similarities within potential matches, allowing it to handle varying lighting conditions, partial occlusions,
    and pose changes while maintaining identity consistency.

\begin{figure}[t]
        \centering
        \includegraphics[width=\textwidth]{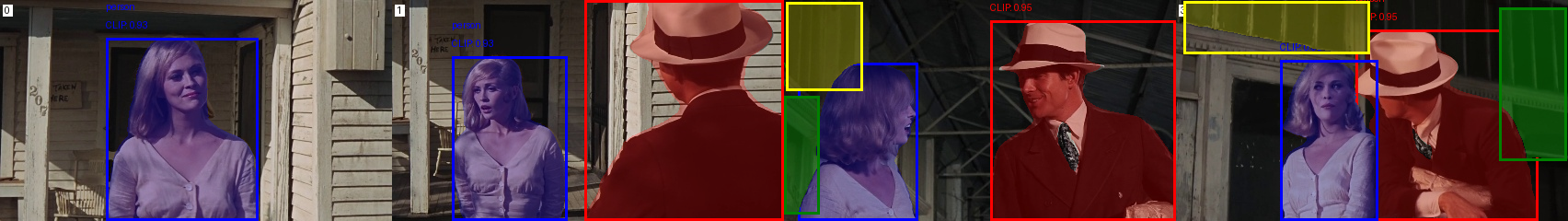}
        \caption{Visualization of cross-frame object re-identification.
        Objects with the same colored bounding boxes across different frames represent instances of the same entity being tracked throughout the sequence.
        The system maintains consistent identity for characters (blue and red boxes) and objects (yellow and green boxes) across multiple frames.}
        \label{fig:cross-frame-detection}
    \end{figure}

    The output of the algorithm is a set of entities, each containing a collection of detections representing the same entity across different frames,
    with assigned global IDs for the entity and IDs for each individual detection within the entity.
    Fig.~\ref{fig:cross-frame-detection} shows our cross-frame object re-identification approach in action.

    \subsection{Structured Scene Analysis}\label{subsec:structured-analysis}
    \glsreset{cot}
    Our structured scene analysis framework explicitly models characters, objects, settings, and narrative structure across frames,
    serving as the backbone for our \gls{cot} reasoning process~\cite{zhang2023multimodal}.
    Our analysis organizes visual content into a hierarchical structure consisting of image-level analyses and cross-frame
    narrative connections.

    For each image in a sequence, our model generates a detailed structured analysis with three key components:
    (1) a character table tracks individuals across frames with consistent identity references while modeling their apparent emotions,
    actions, narrative functions, and spatial location through precise bounding box coordinates.
    Each character entry captures not only physical description but also emotional states and purpose within the narrative,
    enabling storytelling that accounts for character development and motivation;
    (2) an object table documents objects with their functions, interactions, and spatial locations through precise bounding box coordinates,
    maintaining consistent object references across the sequence.
    By explicitly tracking how objects are used or interacted with, the system preserves meaningful object-character
    relationships that might otherwise be lost in translation from visual to textual modalities;
    (3) a setting table categorizes environmental elements using a predefined taxonomy (Location, Environment, Lighting,
    Weather, Time Period, Architecture, Interior Design, Atmosphere, and Background) to provide consistent scene descriptions.
    This approach to setting ensures that environmental context is preserved across frames.

    The system connects these individual image analyses through a narrative structure table that models the story's progression
    through standardized narrative phases: Introduction, Development, Conflict, Turning Point, and Conclusion~\cite{freytag1894freytag}.
    The entire analysis is organized into a hierarchical structure where individual image analyses contain their respective
    character, object, and setting tables, while the story analysis maintains the collection of image analyses
    along with the cross-frame narrative structure.
    By explicitly representing both intra-frame relationships (e.g., character-object interactions) and inter-frame connections
    (e.g., narrative progression and entity re-identification), our system can generate stories that maintain
    coherence at both micro and macro levels.
    This representation provides the foundation for generating grounded stories that maintain coherence across frames.
    The full prompt that guided this process is detailed in Appendix~\ref{subsec:cot-prompt}.

    \subsection{Grounded Story Generation}\label{subsec:grounded-story}
    The structured scene analysis provides a foundation for generating coherent narratives, but transforming this analysis
    into stories requires additional mechanisms for grounding generated text to visual elements.
    Our approach, adapted from GroundCap~\cite{Oliveira2025GroundCapAV} uses a specialized XML-based grounding scheme that
    explicitly links textual elements to visual entities across multiple frames.

    The grounding system employs four types of specialized tags that connect different narrative components to their visual counterparts.
    Image tags (\texttt{<gdi>}) demarcate text segments corresponding to specific frames, ensuring proper temporal alignment.
    Entity reference tags (\texttt{<gdo>}) link both character and object mentions, including pronouns, to their consistent
    identifiers established during cross-frame matching.
    Action tags (\texttt{<gda>}) ground character actions to the individuals performing them, maintaining the connection
    between behaviors and actors.
    Finally, location/landmark tags (\texttt{<gdl>}) ground descriptions of landmarks and background elements to their visual counterparts.

    To ensure data quality, we implemented a validation pipeline that checks:
    (1) entity IDs in the story grounding tags appear in the \gls{cot} analysis;
    (2) bounding box coordinates are properly formatted and within the image dimensions;
    (3) \gls{cot} tables maintain correct structure.
    When errors were detected, generation was repeated until all validation criteria were satisfied.
    This tagging system accommodates single-entity references (e.g., \texttt{<gdo char1>} Sarah \texttt{</gdo>})
    and multi-entity references (e.g., \texttt{<gdo char1 char2>} They \texttt{</gdo>}), enabling grounding of collective pronouns and joint actions.
    The system grounds all references, including pronouns to maintain entity tracking throughout the narrative.

    Our story generation process constructs narratives that balance creative storytelling with visual fidelity.
    The generator \gls{llm}, Qwen-VL 2.5~\cite{qwen25vl} receives the structured analysis generated in the previous step,
    and the original images as inputs.
    It is then asked to create a story while remaining plausible given the visual evidence.
    The generator \gls{llm} first identifies core narrative elements from the structured analysis, focusing on character relationships,
    significant objects, setting details, and narrative phases.
    It then develops a storyline while adhering to the narrative structure identified in the cross-image analysis.
    Throughout this process, the generator \gls{llm} produces text with the grounding tags that link every character and object reference,
    action description, and landmark element to its corresponding visual entity.
    The XML tagging scheme makes the connection between text and visual elements explicit and machine-readable,
    facilitating evaluation and enabling applications tracing narrative elements back to their visual origins.
    The prompt used to generate the story can be found in Appendix~\ref{subsec:story-prompt}.

    \subsection{Post-Processing and Context Length Optimization}\label{subsec:post-processing-and-context-length-optimization}
    To ensure consistency across the dataset, we standardize the format of all entries by parsing the output generated by the \gls{llm} through
    a parser and rewriting it with consistent spacing, line breaks, and structural elements.
    This standardization process guarantees uniformity across the entire dataset, facilitating model training and
    evaluation while maintaining the integrity of the grounding tags and narrative structure.
    To optimize dataset utility across different compute environments, we also provide a dataset loader with selective filtering of entities
    from the \gls{cot} analysis.
    The dataset loader allows removing characters, objects, and settings from the \gls{cot} that are not referenced in the story.
    This selective filtering reduces context length while preserving all narrative-relevant information.
    Additionally, we provide functionality to dynamically truncate stories based on a maximum number of images,
    allowing researchers to train on our dataset with variable context size constraints.

    \subsection{Dataset Statistics}\label{subsec:dataset-statistics}
    To better understand the characteristics of Story Reasoning dataset, we conducted a statistical analysis examining entity persistence,
    grounding effectiveness, and narrative structure.
    The dataset consists of 4,178 stories with an average of 12.44 frames per story, ranging up to 22 frames.
    Stories contain an average of 970.75 words.
    One of the dataset's key strengths is the tracking of entities across multiple frames. 52.39\% of characters and 36.85\% of
    objects appear in at least two frames, with a natural decline in persistence as the frame count increases.
    Each image contains an average of 1.97 characters and 3.38 objects.
    The dataset features extensive entity grounding, with an average of 150.37 entity references per story.
    These references are distributed across character mentions (99.30), object references (11.51),
    setting descriptions (11.59), and action references (27.97).
    Pronouns constitute 45.94\% of all character references.

    Analysis of pronoun grounding reveals that 71.53\% of subject pronouns and 53.26\% of possessive pronouns are grounded to their visual counterparts.
    Notably, gender-specific pronouns (``he''/``she'') demonstrate higher grounding rates (91.79\%/91.82\%) compared to plural pronouns like ``they'' (57.63\% grounded).
    We found that pronouns appearing inside dialogues tend not to be grounded.
    This is especially notable for the pronouns ``I'' and ``we'', which showed the lowest grounding percentages (0.50\% and 4.02\% for ``we'' and ``I'' respectively)
    because these pronouns predominantly appear in character dialogues and the current grounding system does not ground them.
    This observation highlights a specific area for future improvement in the grounding methodology.

    All of the stories in the dataset contain all five narrative phases (Introduction, Development, Conflict, Turning Point, and Conclusion)~\cite{freytag1894freytag},
    chosen for being one of the most widely recognized dramatic structure in storytelling.
    Stories contain an average of 8.44 distinct characters and 27.57 character actions.

    The dataset provides a strong foundation for training models that can generate stories maintaining character and object consistency across multiple frames while grounding narrative elements to visual entities.
    Additional dataset statistics with corresponding charts are available in the appendix.
    The Story Reasoning dataset is publicly available at \url{https://huggingface.co/datasets/daniel3303/StoryReasoning}.

    \section{Qwen Storyteller: The Baseline Model}\label{sec:qwen-storyteller-the-baseline-model}
    To establish a baseline for visual storytelling with cross-frame consistency, we fine-tuned
    Qwen2.5-VL 7B~\cite{qwen25vl} on the Story Reasoning dataset.
    We chose Qwen2.5-VL 7B for its visual reasoning and object detection capabilities.
    Our baseline processes sequences of images directly, performing end-to-end object detection,
    object re-identification, and landmark detection while generating the \gls{cot} representing the scene analysis tables.
    Based on this analysis, it then generates the stories with appropriate grounding tags.
    The entire process from raw images to grounded stories is performed end-to-end by a single model being this
    one of the key innovations with respect to the GroundCap baseline~\cite{Oliveira2025GroundCapAV}.

    We fine-tuned Qwen2.5-VL 7B~\cite{qwen25vl} on the Story Reasoning dataset with filtering of entities from the \gls{cot}
    that are not referenced in the story.
    We tested several training configurations.
    For parameter-efficient training, we used \gls{lora}~\cite{hu2022lora} with (1) rank 512
    with alpha scaling factor 1024, and (2) rank 2048 with alpha scaling factor 4096, both targeting self-attention layers
    (query, key, value, output projections) of the language components.
    We employed full model fine-tuning as a third approach.
    For the \gls{lora} experiments, we used a peak learning rate of $1 \times 10^{-4}$ with batch size 32,
    while the full model fine-tuning used a learning rate of $2 \times 10^{-5}$ with batch size 64.
    All experiments used warmup~\cite{loshchilov2017sgdr} for the first 3\% of
    steps for 4 epochs, AdamW optimizer~\cite{loshchilov2018decoupled} with weight decay 0.01, and bfloat16 precision.
    Training took between 6 and 12 hours on two NVIDIA A100 GPUs (80GB VRAM), depending on the configuration, using
    Liger Kernel~\cite{hsu2024ligerkernelefficienttriton} for efficient memory management.
    The Qwen Storyteller (\gls{lora} Rank 2048) model is publicly available at \url{https://huggingface.co/daniel3303/QwenStoryteller}.

    \section{Evaluation Results}\label{sec:evaluation-results}
    We conducted the evaluation of Qwen Storyteller using both automated metrics and \gls{llm}-based assessments.
    Our evaluation focuses on grounding effectiveness, cross-frame consistency, and hallucination reduction.

    \subsection{Automatic Metrics}
    We first evaluated the grounding capabilities of Qwen Storyteller using standard metrics for object detection (Precision, Recall, F1 Score and \gls{map}~\cite{everingham2010pascal}),
    object re-identification, and language (METEOR~\cite{banerjee2005meteor}).
    We calculate Precision ($P = \frac{TP}{TP+FP}$) as the proportion of objects referenced in the generated story that correctly match ground truth objects.
    Recall ($R = \frac{TP}{TP+FN}$) measures the proportion of ground truth objects that are referenced in the generated story.
    F1 Score ($F1 = 2 \cdot \frac{P \cdot R}{P+R}$) is the harmonic mean of precision and recall, providing a single metric balancing both aspects of grounding performance.

    We adapt \gls{map} metric~\cite{everingham2010pascal} to fit our narrative grounding task, removing the dependency on
    confidence scores that are not available when using \glspl{llm} for detection.
    Our implementation calculates \gls{ap} for each story by examining matched detections between reference and candidate stories.
    While traditional object detection evaluations rely on detector confidence scores, our approach simply orders detections sequentially,
    computing precision and recall at each point.
    Using the 11-point interpolation method, we divide the recall space into 11 equally spaced points from 0 to 1,
    and compute the maximum precision achievable at each recall level.
    The \gls{ap} for each story is calculated as the average of these precision values, and the final \gls{map} is the mean
    across all stories.
    Table~\ref{tab:automatic-metrics} presents these results.

    \begin{table}[ht]
        \centering
        \caption{Automatic evaluation metrics for Qwen Storyteller using different training configurations.
        Precision and Recall are reported for character references (Char), object references (Obj), and combined entities (Total).
        \gls{map} indicates Mean Average Precision across all entities. METEOR (M), ROUGE-L (R) and BLEU-4 (B-4) evaluates linguistic similarity with reference stories.
        \gls{lora} models are trained with rank 512 and 2048, respectively. Full fine-tune indicates full model training.
        Lang fine-tune corresponds to fine-tuning the language model only.
        }
        \label{tab:automatic-metrics}
        \begin{tabular}{l|cccc|ccc|ccc}
            \hline
            & \multicolumn{4}{c|}{Precision} & \multicolumn{3}{c|}{Recall} & \multicolumn{3}{c}{Language} \\
            Model                           & Char          & Obj           & Total         & \gls{map}     & Char          & Obj           & Total         & M             & R             & B-4            \\
            \hline
            \gls{lora}~{\small (Rank 512)}  & \textbf{0.84} & 0.46          & \textbf{0.58} & 0.24          & 0.58          & 0.22          & 0.36 & 0.13 & 0.15 & 0.049 \\
            \gls{lora}~{\small (Rank 2048)} & 0.83          & \textbf{0.46} & 0.57          & \textbf{0.27} & \textbf{0.62} & \textbf{0.25} & \textbf{0.40} & \textbf{0.14} & \textbf{0.16} & \textbf{0.054} \\
            Full fine-tune                  & 0.75          & 0.45          & 0.56          & 0.16          & 0.37          & 0.13          & 0.23          & 0.12          & 0.14          & 0.038          \\
            Lang fine-tune                  & 0.72          & 0.45          & 0.54          & 0.14          & 0.36          & 0.13          & 0.22          & 0.13          & 0.15          & 0.044          \\
            \hline
        \end{tabular}
    \end{table}

    \subsection{LLM-based Evaluation}
    For a more comprehensive assessment of story quality, we employed four state-of-the-art \glspl{llm}:
    Claude 3.7 Sonnet~\cite{anthropic2025}, ChatGPT-4o~\cite{openai2024gpt4o},
    Gemini Flash 2.5~\cite{google2025gemini} and Qwen 2.5-VL 72B~\cite{qwen25vl}.
    We generated stories using both the fine-tuned Qwen Storyteller using \gls{lora} rank 2048
    and the non-fine-tuned Qwen2.5-VL 7B model on the same image sequences.
    Each \gls{llm} evaluated stories using a 5-point Likert scale (1: Poor, 5: Excellent) for Description Accuracy,
    while hallucinations were quantified as absolute counts across different categories.

    Following the hallucination taxonomy proposed by Huang et al.~\citep{Liu2024ASO}, we categorized hallucinations into four distinct
    types: Object Hallucination, Attribute Hallucination, Relationship Hallucination, and Environmental Detail Hallucination.
    Table~\ref{tab:llm-eval-results} presents the \gls{llm} evaluation scores.
    Qwen Storyteller achieved higher description accuracy than the base model (2.76 vs 2.69), significantly improved creativity (3.38 vs 2.58, +31.0\%), while
    reducing the total hallucinations from 4.06 to 3.56 (-12.3\%) on average, contributing to a better story generation process when compared
    with the non-fine-tuned model.
    Further context regarding the change in the average number of hallucinations per story is provided in the Appendix~\ref{sec:hallucination-analysis}.

    \begin{table}[ht]
        \centering
        \caption{LLM average ratings for description accuracy (scale: 1-5), creativity (scale: 1-5), and hallucination counts by category (lower is better). Qwen Storyteller corresponds to the fine-tuned model using \gls{lora} rank 2048. Rel. stands for Relationship and Env. for Environmental.}
        \label{tab:llm-eval-results}
        \begin{tabular}{l|l|l|lllll}
            \hline
            Model & Description & Creativity & \multicolumn{5}{c}{Hallucination Count} \\
            & Accuracy &      & Object & Attribute & Rel. & Env. & Total \\
            \hline
            Qwen2.5-VL 7B    & 2.69     & 2.58 & 1.84   & 1.47      & 0.60 & 0.15 & 4.06  \\
            Qwen Storyteller & 2.76     & 3.38 & 1.54   & 1.28      & 0.63 & 0.11 & 3.56  \\
            \hline
        \end{tabular}
    \end{table}

    In the context of our visual storytelling system, the hallucination types are defined as follows:
    (1) Object Hallucination involves describing characters or objects not present in any frame;
    (2) Attribute Hallucination refers to incorrect descriptions of visual properties such as colors, sizes, or
    physical characteristics;
    (3) Relationship (Rel.) Hallucination involves incorrect spatial or interactive relationships between characters and objects,
    these include incorrect actions or interactions between entities;
    and (4) Environmental (Env.) Detail Hallucination refers to fabricated background elements, settings, or contextual
    details not supported by visual evidence.

    \subsection{Inter-evaluator Agreement}
    To assess the reliability of our \gls{llm}-based evaluations, we measured agreement between the three \gls{llm}
    evaluators using Krippendorff's alpha coefficient~\cite{krippendorff1970estimating}.

    Table~\ref{tab:alpha-coefficients} shows fair agreement across all criteria ($\alpha > 0.14$),
    with relatively higher consensus on hallucination detection (0.34) and character consistency (0.35).
    This agreement suggests that \gls{llm} evaluations provide assessments of visual storytelling quality, though variability exists between evaluators, particularly for description accuracy (0.14).

    \begin{table}[ht]
        \centering
        \caption{Krippendorff's Alpha Coefficients for Inter-evaluator Agreement}
        \label{tab:alpha-coefficients}
        \begin{tabular}{lllllll}
            \hline
            & Hallucination & Character   & Grounding & Object    & Description & Creativity \\
            & Rate          & Consistency & Recall    & Precision & Accuracy    &            \\
            \hline
            Alpha & 0.34          & 0.35        & 0.24      & 0.26      & 0.14        & 0.20       \\
            \hline
        \end{tabular}
    \end{table}

    Overall, our evaluation results demonstrate that Qwen Storyteller outperforms the base model in visual storytelling quality,
    with higher description accuracy (2.76 vs 2.69), significantly improved creativity (3.38 vs 2.58, +31.0\%), and reduced total hallucinations (3.56 vs 4.06 total hallucinations per story).

    \section{Limitations}\label{sec:limitations}
    Despite Qwen Storyteller's advances, several limitations remain.
    The model's re-identification algorithm relies primarily on object appearance rather than considering
    overall image context, which can lead to confusion when similar-looking objects or persons (such as twins)
    appear across frames or within the same frame.
    Our movie-derived dataset introduces biases from cinematic composition and narrative structure that may not
    generalize to candid or unstructured visual sequences.
    First-person pronouns have low grounding rates as they primarily appear in character dialogues which our system fails to ground.
    Additionally, our evaluation methods have inherent limitations as \gls{llm}-based assessments may contain biases and
    automatic language metrics potentially penalize creative narratives than deviate from ground truth.
    Our conclusions are limited to the narrative structure represented by the stories in our dataset.
    Further work is needed to address other narrative structures.

    \section{Conclusions}\label{sec:conclusions}
    We introduced Story Reasoning, a dataset containing 4,178 stories derived from 52,016 movie images, with both structured
    scene analyses and grounded stories that maintain character and object consistency across frames.
    The dataset features cross-frame object re-identification using visual similarity and face recognition,
    chain-of-thought reasoning for narrative modeling, and a grounding scheme linking textual elements to visual entities across multiple frames.
    We established baseline performance by fine-tuning Qwen2.5-VL 7B, creating Qwen Storyteller, which performs end-to-end object detection,
    re-identification, and landmark detection while maintaining consistent object references throughout generated narratives.
    Evaluation demonstrates a reduction from 4.06 to 3.56 (-12.3\%) hallucinations per story and an improvement in creativity from 2.58 to 3.38 (+31.0\%) compared to a non-fine-tuned model.

    \section*{Acknowledgments}\label{sec:acknowledgments}
    Daniel Oliveira is supported by a scholarship granted by Fundação para a Ciência e Tecnologia (FCT), with reference 2021.06750.BD. Additionally, this work was supported by Portuguese national funds through FCT, with reference UIDB/50021/2020.

    \bibliographystyle{elsarticle-num}
    \bibliography{bibliography}

    \appendix

    \section{Dataset Analysis}\label{sec:dataset-analysis}
    This section presents detailed analyses and statistics of the Story Reasoning dataset to provide insight into its composition and characteristics.
    These analyses complement the overview provided in Section~\ref{sec:the-story-reasoning-dataset} by offering quantitative evidence of the dataset's properties.

    \subsection{Frame Distribution Analysis}\label{subsec:frame-distribution}
\begin{figure}[htbp]
        \centering
        \includegraphics[width=0.7\textwidth]{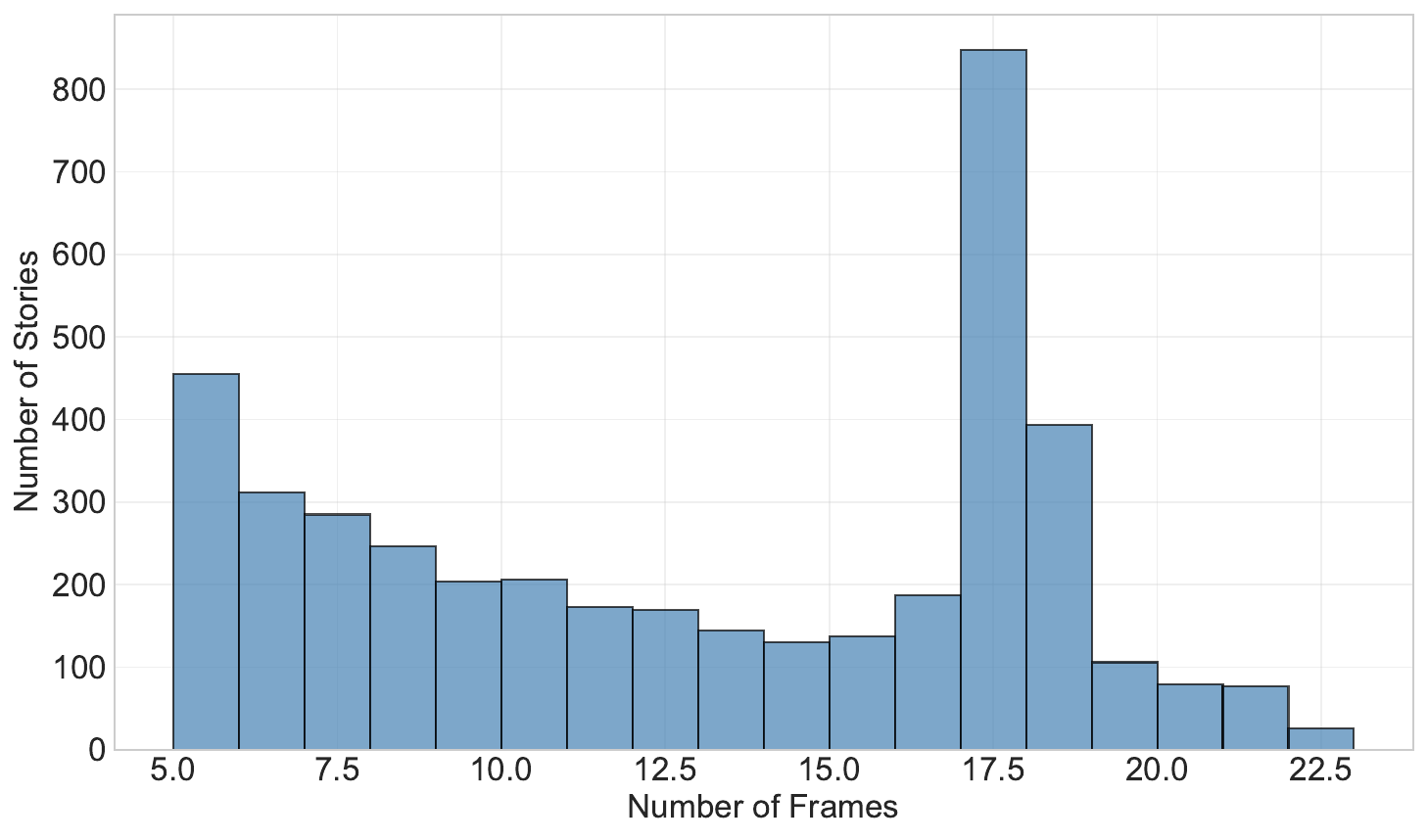}
        \caption{Distribution of frame counts across stories in the Story Reasoning dataset.}
        \label{fig:frame-distribution}
    \end{figure}

    Fig.~\ref{fig:frame-distribution} shows the distribution of frame counts across the 4,178 stories in the Story Reasoning dataset.
    There are a mean of 12.44 frames per story.

    \subsection{Entity Persistence Analysis}\label{subsec:entity-persistence}
\begin{figure}[htbp]
        \centering
        \includegraphics[width=0.7\textwidth]{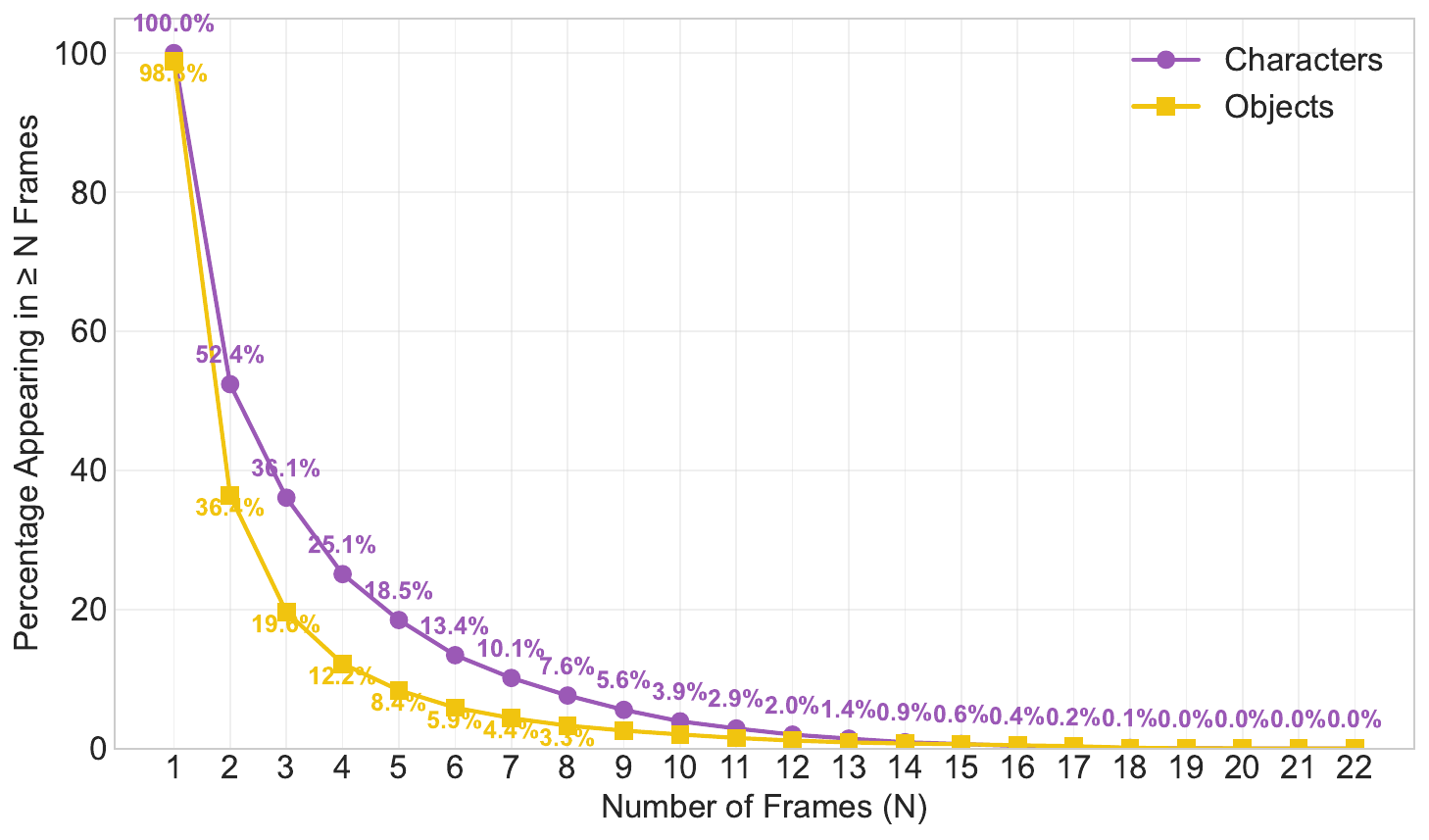}
        \caption{Entity persistence across multiple frames. The graph shows the percentage of characters and objects that appear in at least N frames, demonstrating how entity presence decreases as frame count increases.}
        \label{fig:cross-frame-consistency}
    \end{figure}

    Fig.~\ref{fig:cross-frame-consistency} shows the persistence of characters and objects across frames in our dataset.
    52.4\% of characters appear in at least two frames, compared to 36.4\% of objects, demonstrating that characters
    tend to persist longer than objects in story sequences.
    18.5\% of characters and 8.4\% of objects appear in at least 5 frames which corresponds to the minimum number of frames of a story.

    This differential persistence reflects typical narrative structure, where characters drive stories while objects play supporting roles.
    The declining curves show how visual narratives naturally progress, with fewer entities remaining visible across multiple frames.

    \subsection{Grounding Reference Distribution}\label{subsec:grounding-reference}
\begin{figure}[htbp]
        \centering
        \includegraphics[width=0.7\textwidth]{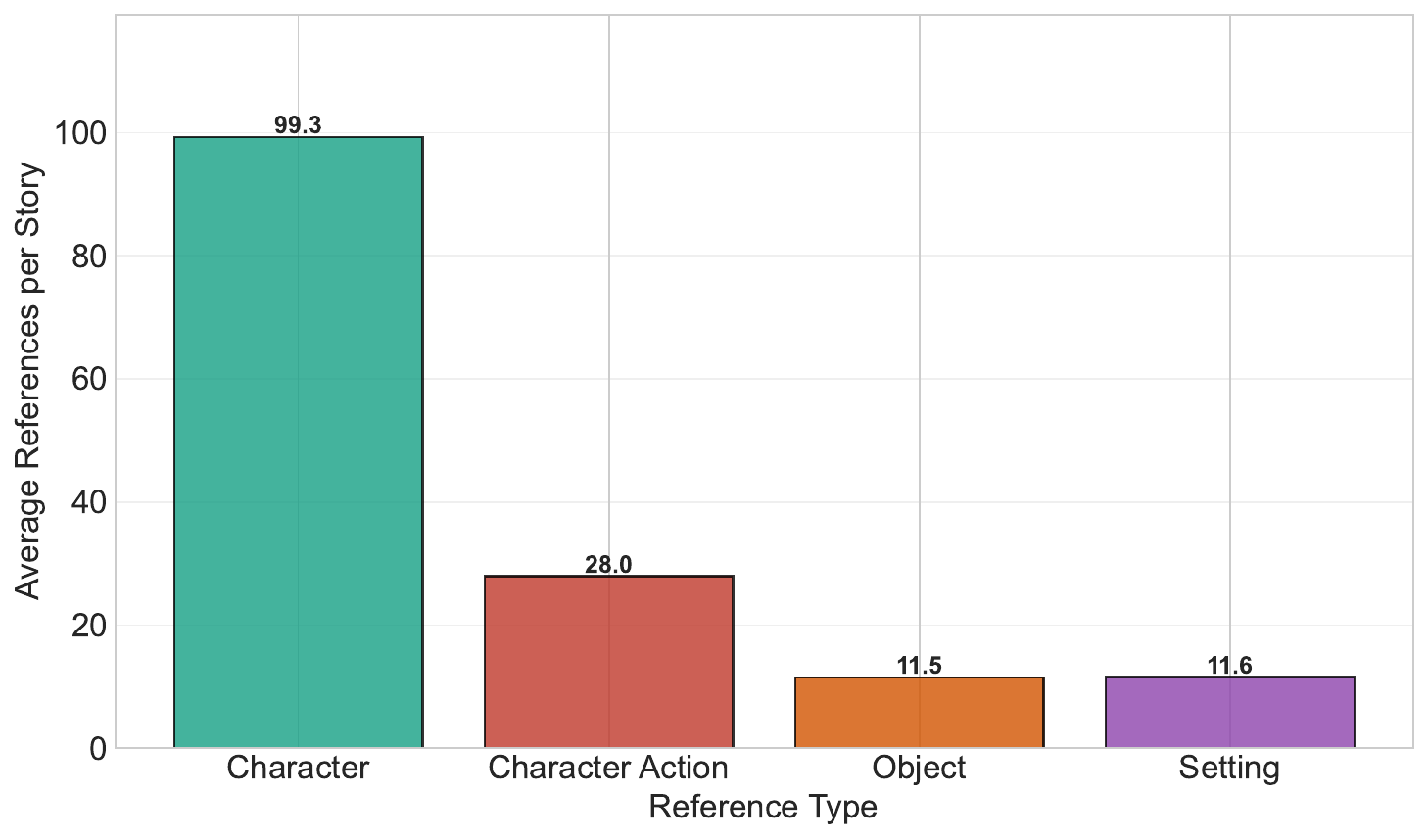}
        \caption{Average number of grounding references by type of entity per story.}
        \label{fig:grounding-stats}
    \end{figure}

    Fig.~\ref{fig:grounding-stats} presents the average number of grounding references by entity type across all stories in the dataset.
    Character references dominate the distribution with an average of 99.3 references per story.
    This reflects the character-centric nature of narrative storytelling, where people drive the plot through their actions and interactions.

    Character actions account for the second-highest category with 28.0 references per story, linking characters to their behaviors
    and establishing causal relationships within the narrative.
    Object references (11.5 per story) and setting references (11.6 per story) appear in similar proportions, these provide contextual
    elements that support the narrative.

    \subsection{Narrative Structure Analysis}\label{subsec:narrative-structure}
\begin{figure}[htbp]
        \centering
        \includegraphics[width=0.7\textwidth]{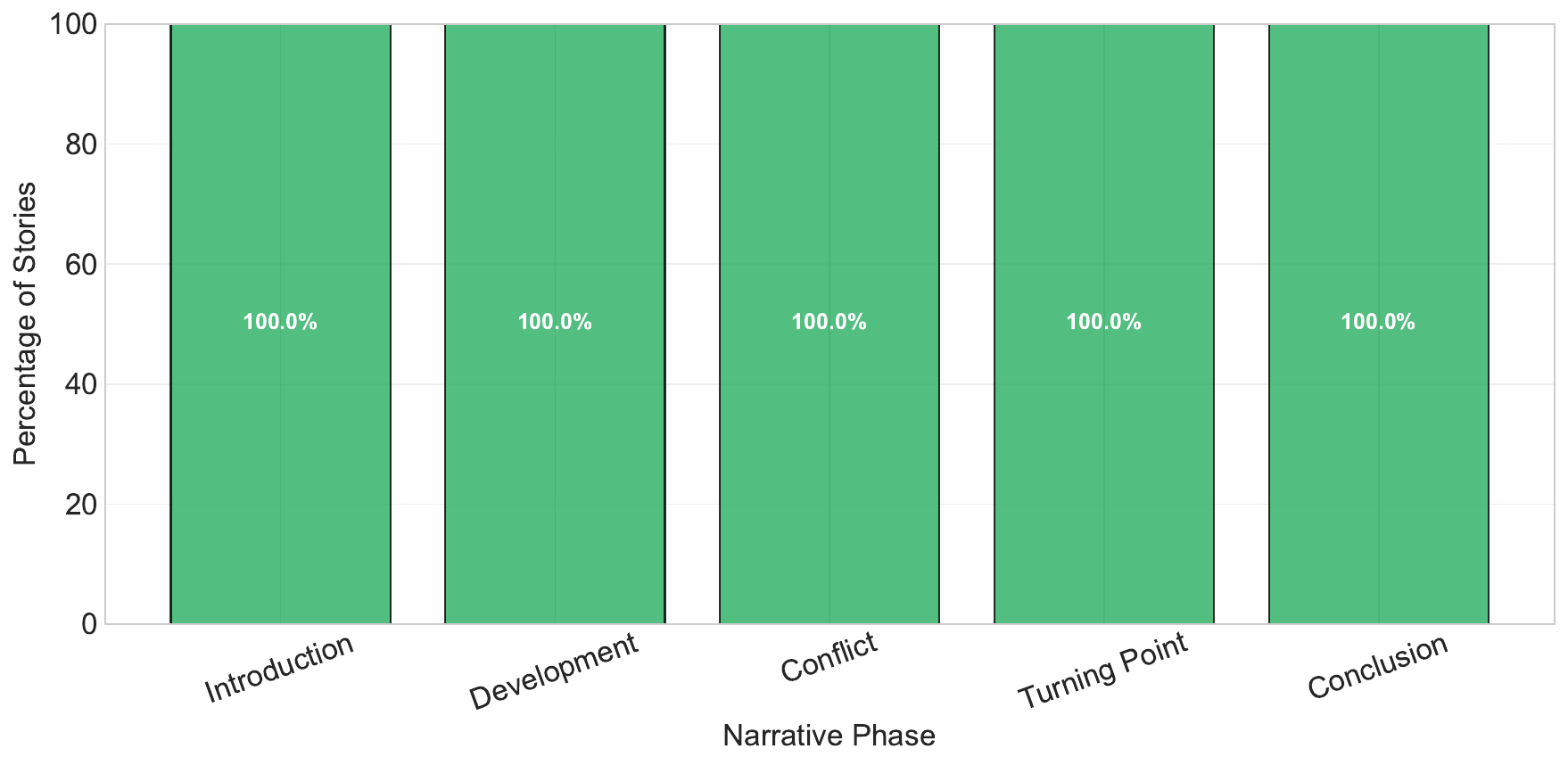}
        \caption{Distribution of narrative phases across stories, showing the percentage of stories containing each narrative phase.}
        \label{fig:narrative-phase-distribution}
    \end{figure}

    Fig.~\ref{fig:narrative-phase-distribution} shows the distribution of narrative phases across stories in our dataset.
    All stories contain each of the five standard narrative phases: Introduction, Development, Conflict, Turning Point, and Conclusion.
    This complete coverage of narrative phases was an intentional design feature of our dataset creation process.

    \subsection{Entity Density Analysis}\label{subsec:entity-density}
\begin{figure}[htbp]
        \centering
        \includegraphics[width=\textwidth]{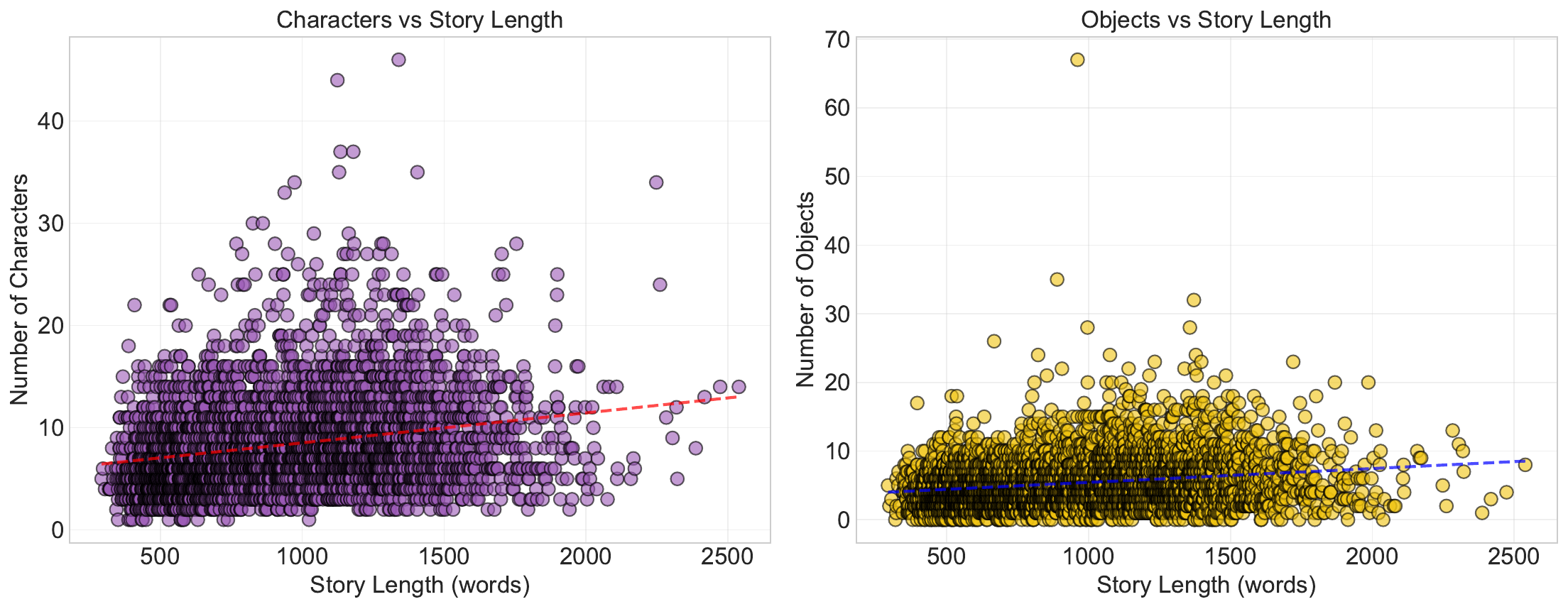}
        \caption{Relationship between story length (in words) and the number of characters and objects, showing positive trend lines for both entity types with story length.}
        \label{fig:entities-by-story-length}
    \end{figure}

    Fig.~\ref{fig:entities-by-story-length} shows the relationship between story length and the number of entities present.
    The left plot displays character count versus story length, while the right plot shows object count versus story length.
    Both plots include trend lines indicating a weak positive relationship.

    The character plot shows a slightly steeper trend line than the object plot, suggesting that increasing story length more reliably
    corresponds with higher character counts than object counts.
    There is an average of 8.44 characters and 5.36 objects per story.
    These trends show that as stories grow in length, they tend to incorporate additional entities rather than simply elaborating on a fixed set of characters and objects.

    \subsection{Grounding Density Across Narrative Phases}\label{subsec:grounding-density}
\begin{figure}[htbp]
        \centering
        \includegraphics[width=0.7\textwidth]{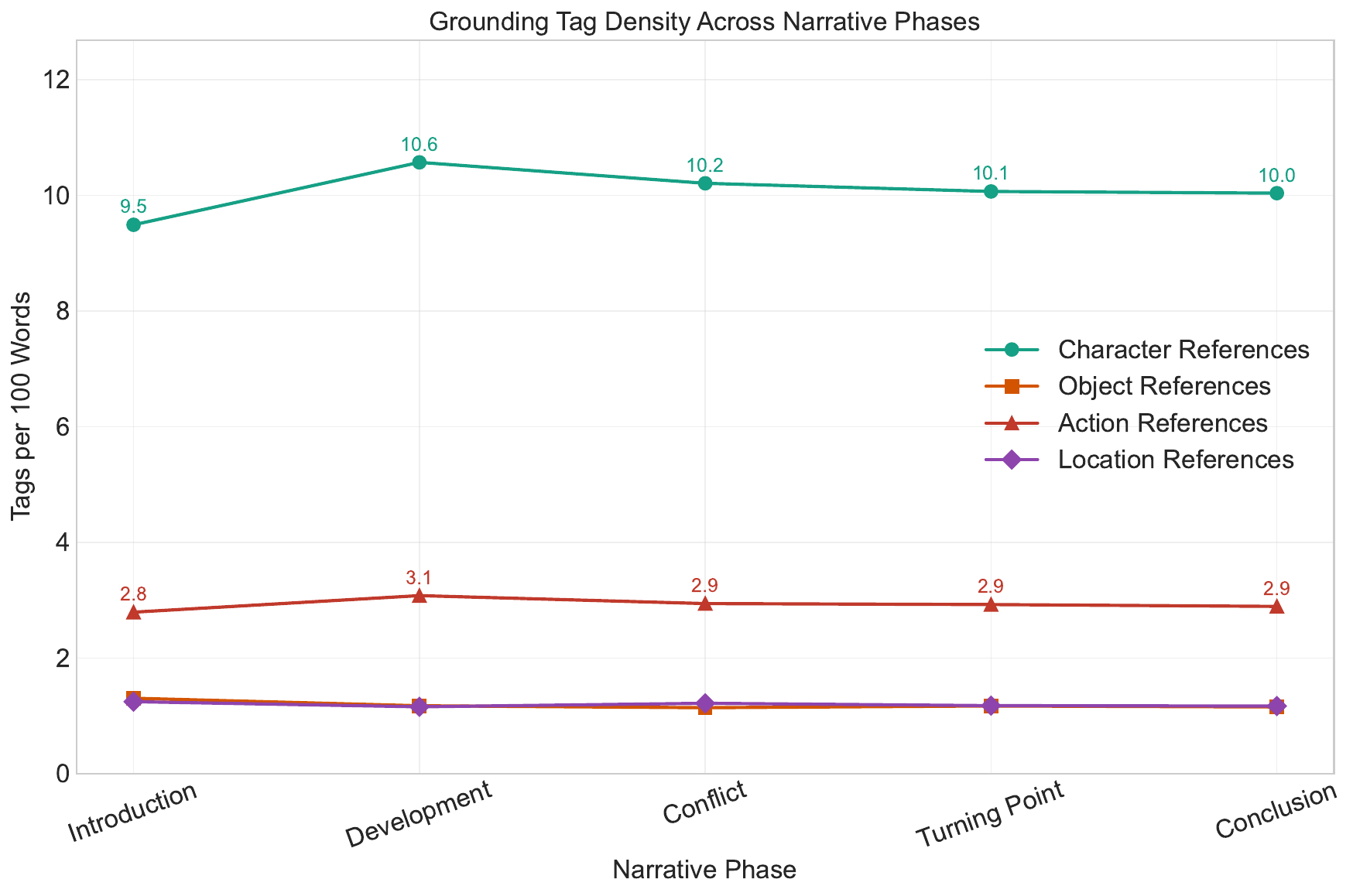}
        \caption{Distribution of grounding references across narrative phases, showing the density of character, object, and setting references in each phase of the story structure.}
        \label{fig:grounding-density-by-phase}
    \end{figure}

    Fig.~\ref{fig:grounding-density-by-phase} shows the distribution of grounding tag density across the five narrative phases.
    Character references dominate across all phases, with the highest density occurring in the Development phase (10.6 tags per 100 words)
    and the lowest in the Introduction phase (9.5 tags per 100 words).

    Action references maintain a relatively consistent density throughout the narrative structure, with a slight peak in the
    Development phase (3.1 tags per 100 words) before stabilizing at 2.9 tags per 100 words for the remaining phases.
    Object and location references show minimal variation across phases, maintaining the lowest densities among all reference types.

    \subsection{Pronoun Grounding Analysis}\label{subsec:pronoun-grounding}
\begin{figure}[htbp]
        \centering
        \includegraphics[width=0.7\textwidth]{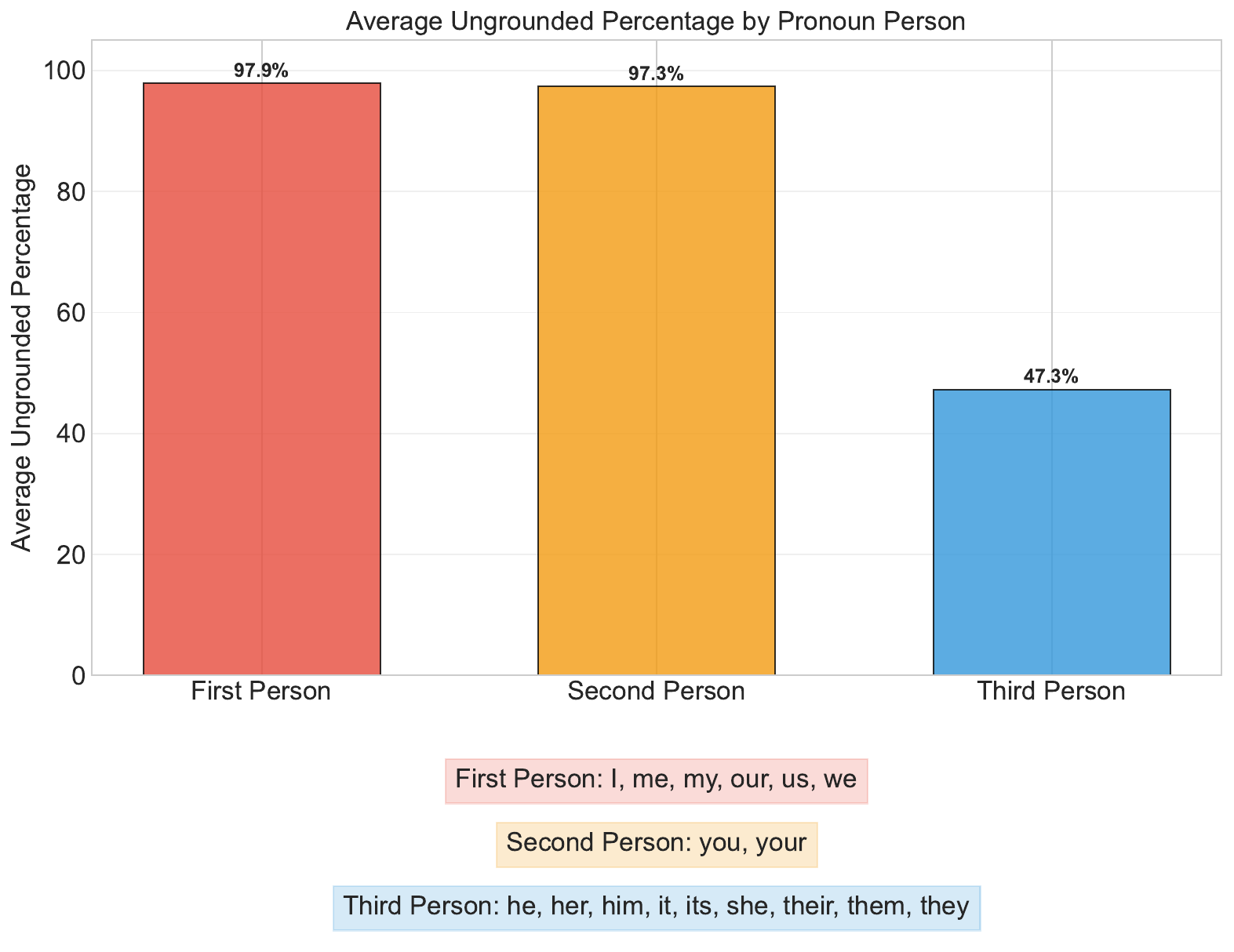}
        \caption{Comparison of pronoun usage between first, second, and third person across stories. }
        \label{fig:pronoun-person-comparison}
    \end{figure}

    Fig.~\ref{fig:pronoun-person-comparison} shows the percentage of ungrounded pronouns by grammatical person.
    First-person pronouns (I, me, my, our, us, we) have the highest ungrounded rate at 97.9\%, followed by second-person pronouns (you, your) at 97.3\%. Third-person pronouns (he, she, they, etc.) show a significantly lower ungrounded rate of 47.3%.
    This disparity exists because our system systematically fails to ground character dialogues and first and second-person pronouns appear mostly in dialogues.
    The 47.3\% ungrounded rate for third-person pronouns represents cases where our system ``forgets'' to ground the pronoun, specially when the pronoun appears after a proper noun that has already been grounded.

\begin{figure}[htbp]
        \centering
        \includegraphics[width=\textwidth]{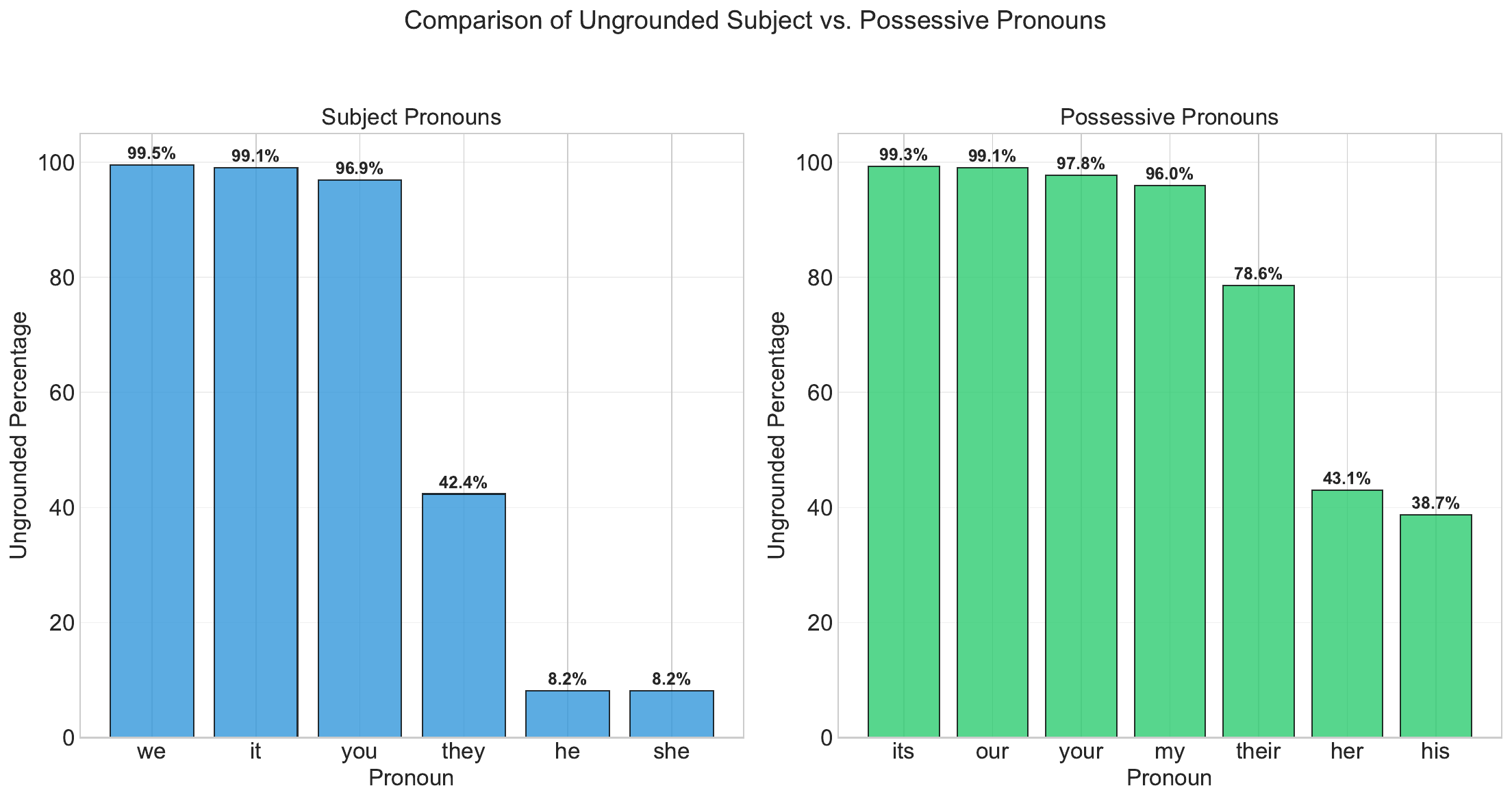}
        \caption{Grounding rates between subject pronouns and possessive pronouns.}
        \label{fig:subject-vs-possessive-pronouns}
    \end{figure}

    Fig.~\ref{fig:subject-vs-possessive-pronouns} provides a more detailed breakdown of ungrounded rates by specific pronouns,
    comparing subject and possessive forms.
    Among subject pronouns, first and second-person forms show extremely high ungrounded rates (``we´´ at 99.5\%, ``you'' at 96.9\%),
    while third-person gendered pronouns have low ungrounded rates (``he'' and ``she'' both at 8.2\%).
    The pronoun ``they'' shows a moderate ungrounded rate of 42.4\%, reflecting its use in both dialogue and narrative contexts.

    Possessive pronouns follow a similar pattern but with generally higher ungrounded rates.
    First and second-person possessives remain largely ungrounded (``our'' at 99.1\%, ``your'' at 97.8\%),
    while third-person gendered possessives show better grounding though still higher ungrounded rates than their subject counterparts
    (``his'' at 38.7\%, ``her'' at 43.1\%). The possessive ``their'' has a higher ungrounded rate (78.6\%) compared to its subject counterpart
    ``they''.

    These statistics further confirm that our system grounds third-person singular pronouns in narrative descriptions while struggling
    with first and second-person pronouns in dialogue contexts.

    \subsection{Narrative Sentiment Analysis}\label{subsec:narrative-sentiment}
\begin{figure}[htbp]
        \centering
        \includegraphics[width=\textwidth]{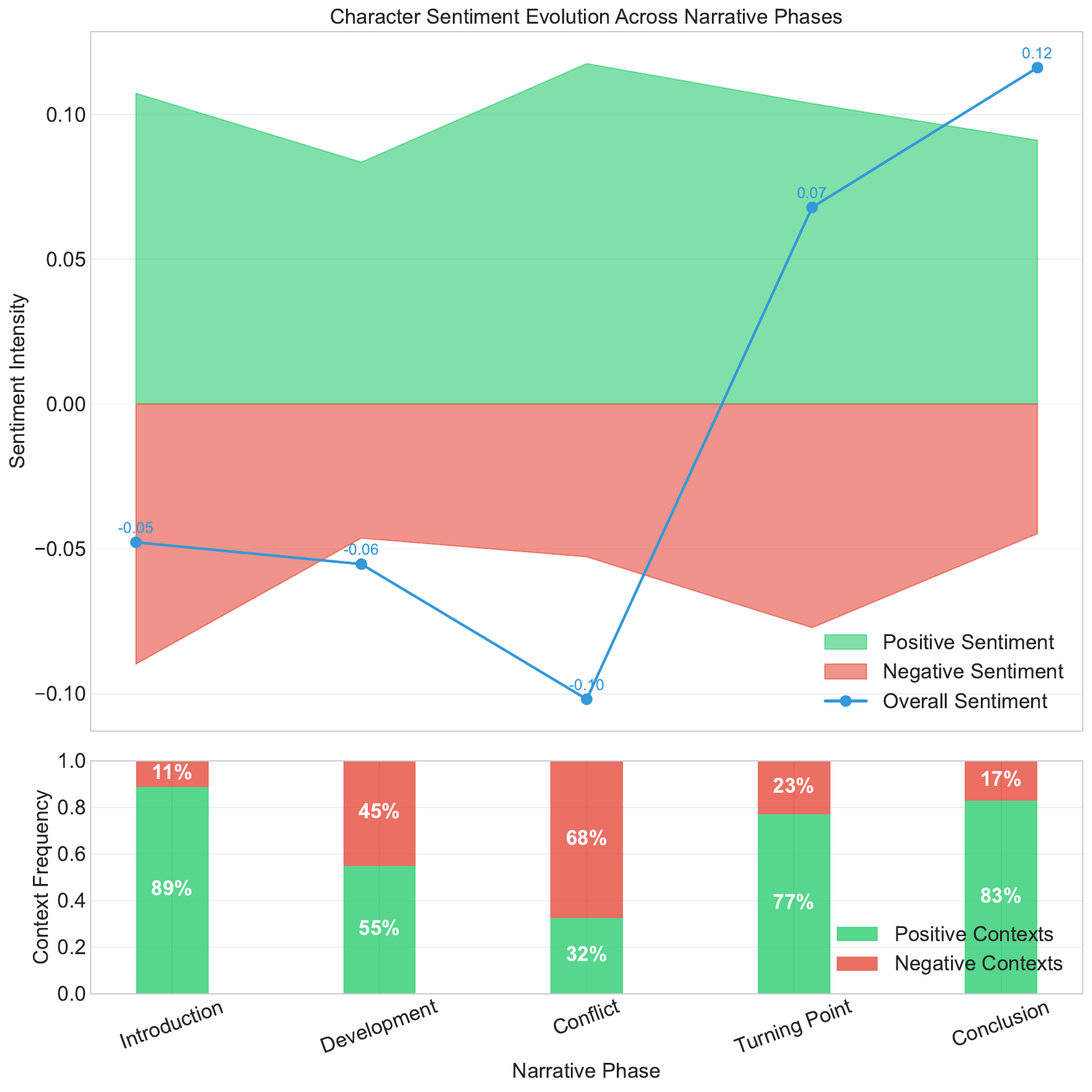}
        \caption{Character sentiment evolution across narrative phases in the StoryReasoning dataset.
        The top panel displays sentiment intensity: the blue line shows overall compound sentiment scores for each phase's complete text,
            while green and red areas represent the average intensity of positive and negative sentiments in character-specific contexts.
            The bottom panel shows the frequency distribution of positive versus negative character contexts across phases.
        }
        \label{fig:entity-sentiment-evolution}
    \end{figure}

    Fig.~\ref{fig:entity-sentiment-evolution} shows the evolution of character sentiment across the five narrative phases in our dataset.
    The analysis, performed using VADER~\cite{hutto2014vader} sentiment analysis on character-specific text segments, reveals an emotional
    arc that aligns with classical dramatic structure.

    The sentiment analysis methodology operates at two distinct levels to capture both the overall narrative sentiment and
    character-specific emotional contexts.
    The blue line in Fig.~\ref{fig:entity-sentiment-evolution} represents the compound sentiment score calculated from the complete text of
    each narrative phase across all stories.
    The positive and negative sentiment areas (green and red) and frequency distributions are derived from localized character contexts only.
    For each character mention in the text (identified by <gdo> tags and identifiers started by char), we extract a surrounding window of
    text (100 characters before and after the mention) and calculate the sentiment of this specific context.
    This character-centric approach allows us to isolate and analyze the emotional atmosphere specifically surrounding characters
    as they navigate through the narrative.
    This dual analysis reveals that while the overall narrative may have a particular emotional tone
    (represented by the blue line), the immediate context around character mentions may demonstrate different patterns
    (shown in the colored areas and frequency bars).
    This is particularly evident in the Introduction phase, where despite 89\% of character contexts being positive, the overall
    narrative sentiment is slightly negative (-0.05), suggesting subtle emotional undercurrents in the broader text that
    foreshadow upcoming conflicts.

    The overall sentiment trajectory begins slightly negative in the Introduction phase (-0.05) and Development phase (-0.06),
    before reaching its lowest point during the Conflict phase (-0.10).
    This corresponds to the narrative tension found in the conflict section of storytelling.
    Following this, sentiment rises through the Turning Point phase (0.07) and continues upward to reach its peak in the Conclusion phase (0.12).
    The frequency distribution of positive versus negative character contexts follows the same pattern as the overall sentiment trajectory.
    Introduction begins with predominantly positive contexts (89\% positive), followed by a gradual shift during
    Development (55\% positive).
    During Conflict, we observe a significant inversion where negative contexts dominate (68\% negative),
    before returning to predominantly positive during Turning Point (77\% positive) and Conclusion (83\% positive).
    This pattern demonstrates the dataset's adherence to traditional narrative structure, with emotional valence following the expected trajectory:
    setup, complication, lowest point, resolution, and positive conclusion.

    \newpage
    \clearpage
    \pagebreak[4]

    \section{Hallucination Analysis}\label{sec:hallucination-analysis}
\begin{figure}[htbp]
        \centering
        \begin{subfigure}[b]{\textwidth}
            \centering
            \includegraphics[width=0.7\textwidth]{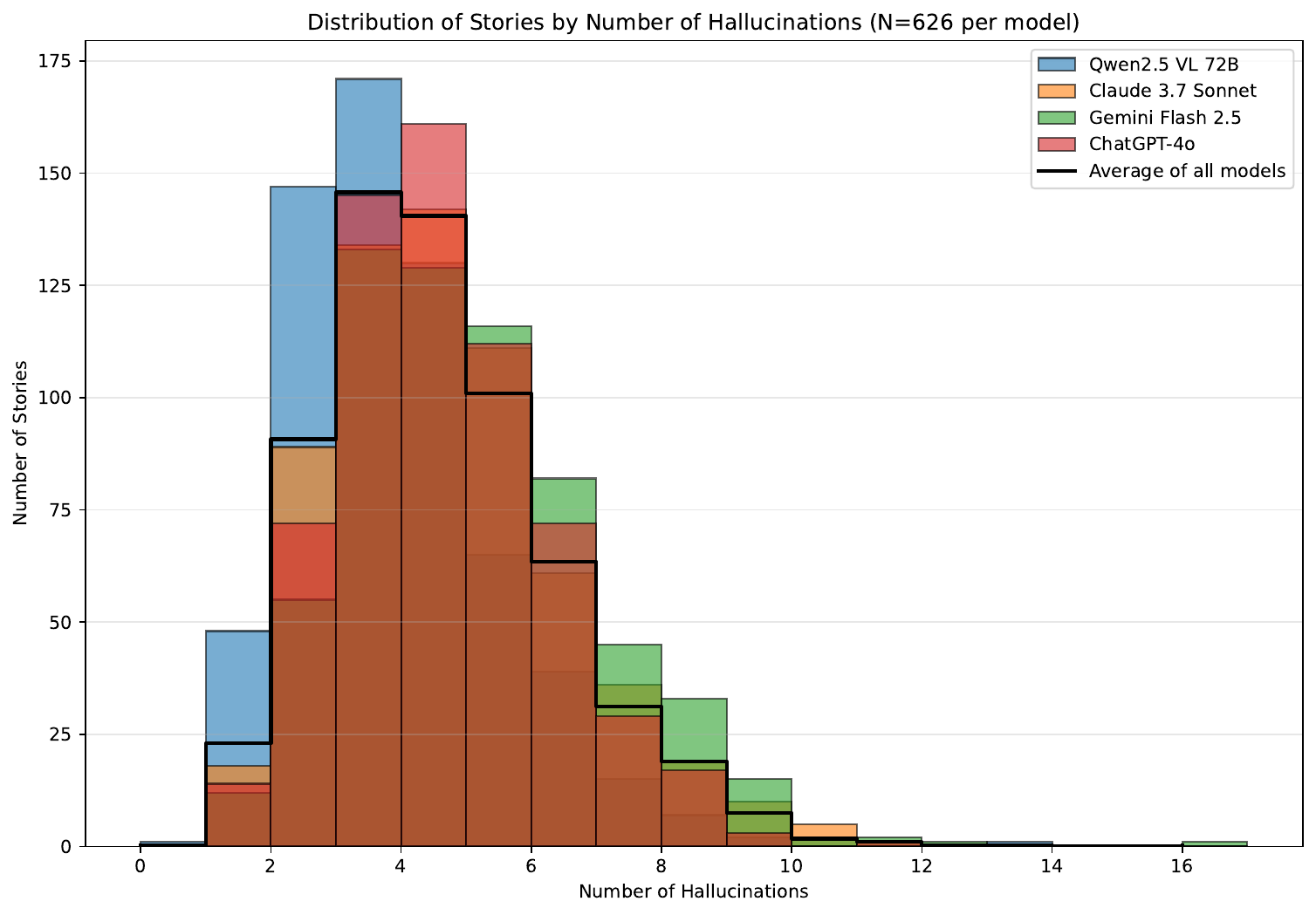}
            \caption{Qwen 2.5-VL 7B model (average: 4.06 hallucinations per story)}
            \label{fig:hallucination-distribution-qwen7b}
        \end{subfigure}

        \vspace{0.5cm}

        \begin{subfigure}[b]{\textwidth}
            \centering
            \includegraphics[width=0.7\textwidth]{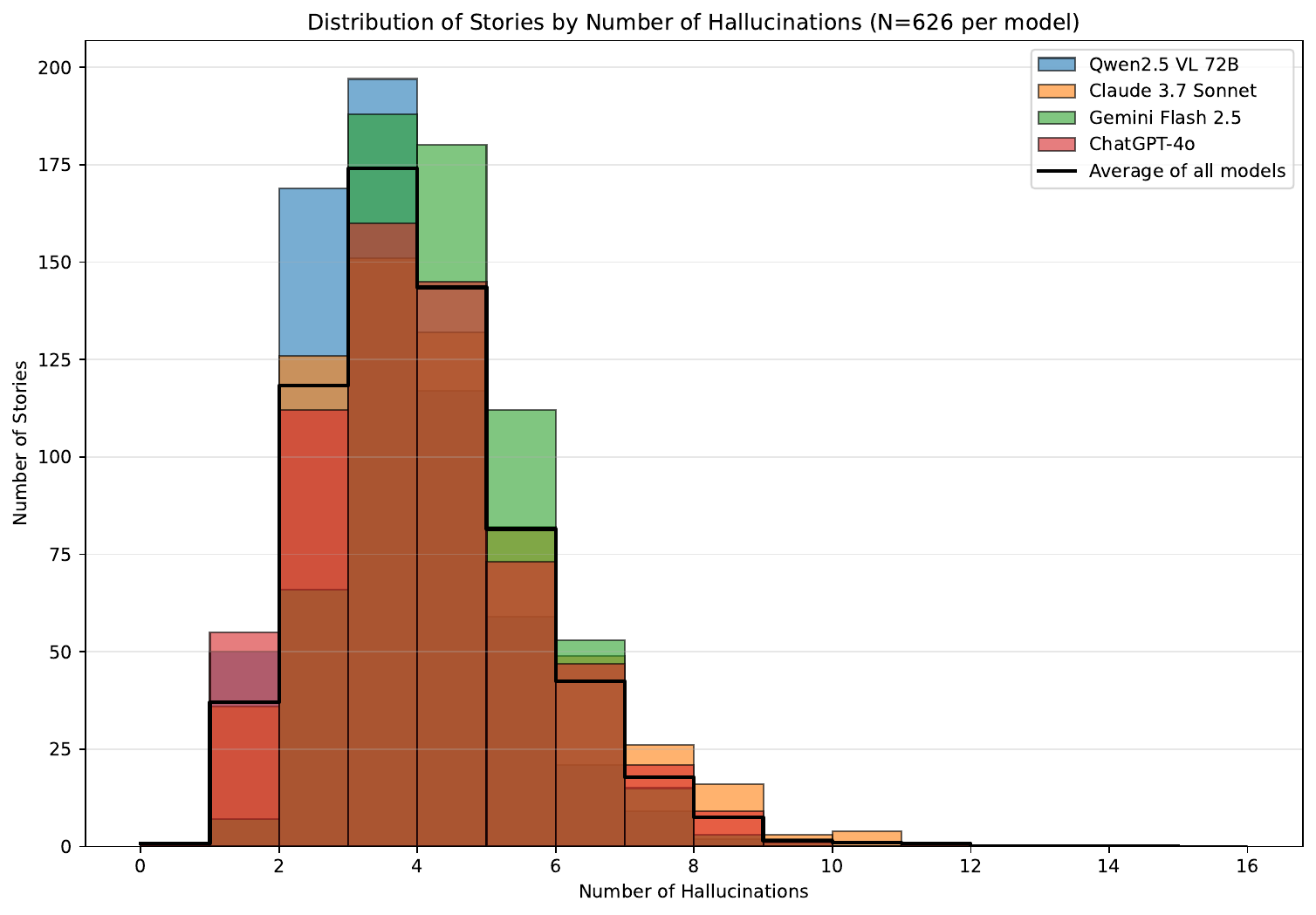}
            \caption{Fine-tuned Qwen Storyteller model (average: 3.56 hallucinations per story)}
            \label{fig:hallucination-distribution-storyteller}
        \end{subfigure}
        \caption{Distribution of hallucinations in stories generated by Qwen2.5-VL 7B and fine-tuned model using \gls{lora} rank 2048.
        The histograms show the number of stories (y-axis) containing specific counts of hallucinations (x-axis), demonstrating a 12.3\% reduction in average hallucinations after fine-tuning.}
        \label{fig:hallucination-distribution}
    \end{figure}

    To better understand the impact of our fine-tuning approach on hallucination reduction, we analyzed the distribution of
    hallucinations in generated stories both before and after fine-tuning.
    Fig.~\ref{fig:hallucination-distribution-qwen7b} shows the distributions for the Qwen2.5-VL 7B model and
    Fig.~\ref{fig:hallucination-distribution-storyteller} shows the distributions for our fine-tuned Qwen Storyteller
    model using \gls{lora} rank 2048.

    As shown in the figure, fine-tuning shifted the distribution toward lower hallucination counts.
    The Qwen Storyteller model produces more stories with 2-3 hallucinations and fewer stories with 5+ hallucinations compared to the base model.
    This shift represents a 12.3\% reduction in average hallucinations from 4.06 to 3.56 on average per story.

    \newpage
    \clearpage
    \pagebreak[4]

    \section{Interactive Visualization Interface}\label{sec:interactive-visualization}
\begin{figure}[htbp]
        \centering
        \includegraphics[width=\textwidth]{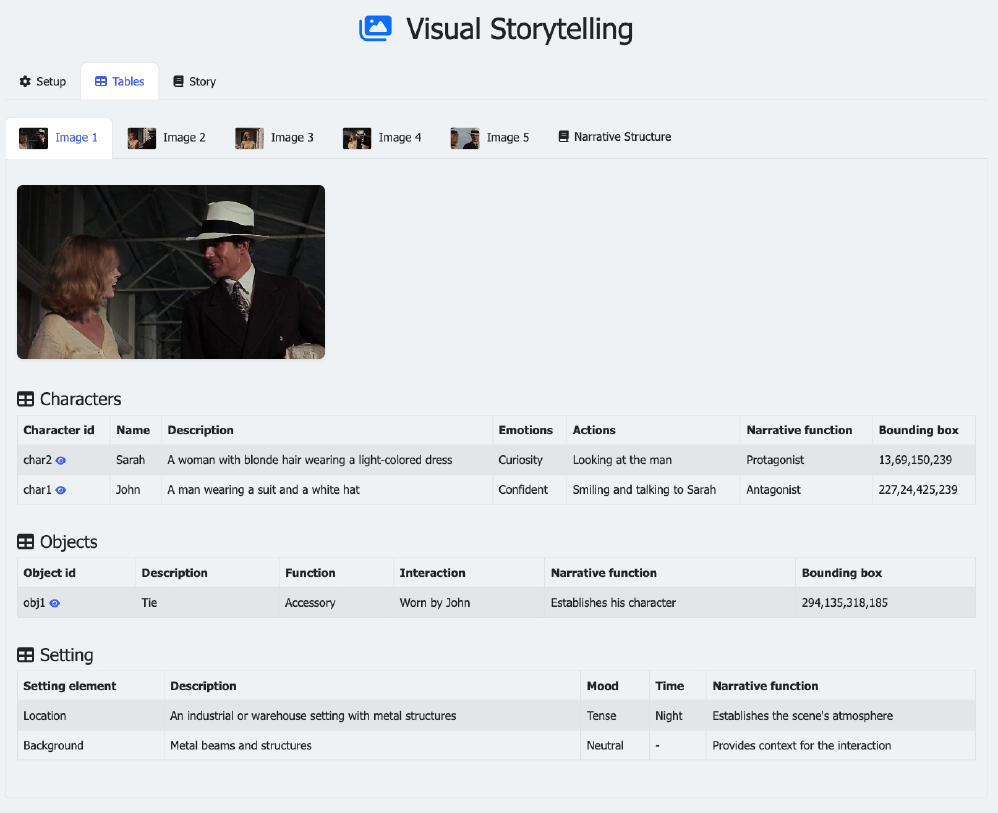}
        \caption{Interface for visualizing the \gls{cot} analysis tables in real-time. The tabbed interface organizes data by image, showing structured information about characters, objects, and setting elements with their properties and bounding box coordinates. Users can navigate between frames to see the complete scene analysis.}
        \label{fig:cot-interface}
    \end{figure}

    To facilitate usage of QwenStoryteller, we developed a web-based interface that allows users
    to generate and visualize stories in real-time.
    The interface code is available at \url{https://github.com/daniel3303/StoryReasoning/blob/master/story_reasoning/visualization/web_interface.html}.
    Fig.~\ref{fig:cot-interface} shows the tabular view of the \gls{cot} analysis, displaying structured information about characters,
    objects, and setting elements.
    The interface organizes data by image, with each tab presenting detailed tables for the corresponding frame.

\begin{figure}[htbp]
        \centering
        \includegraphics[width=\textwidth]{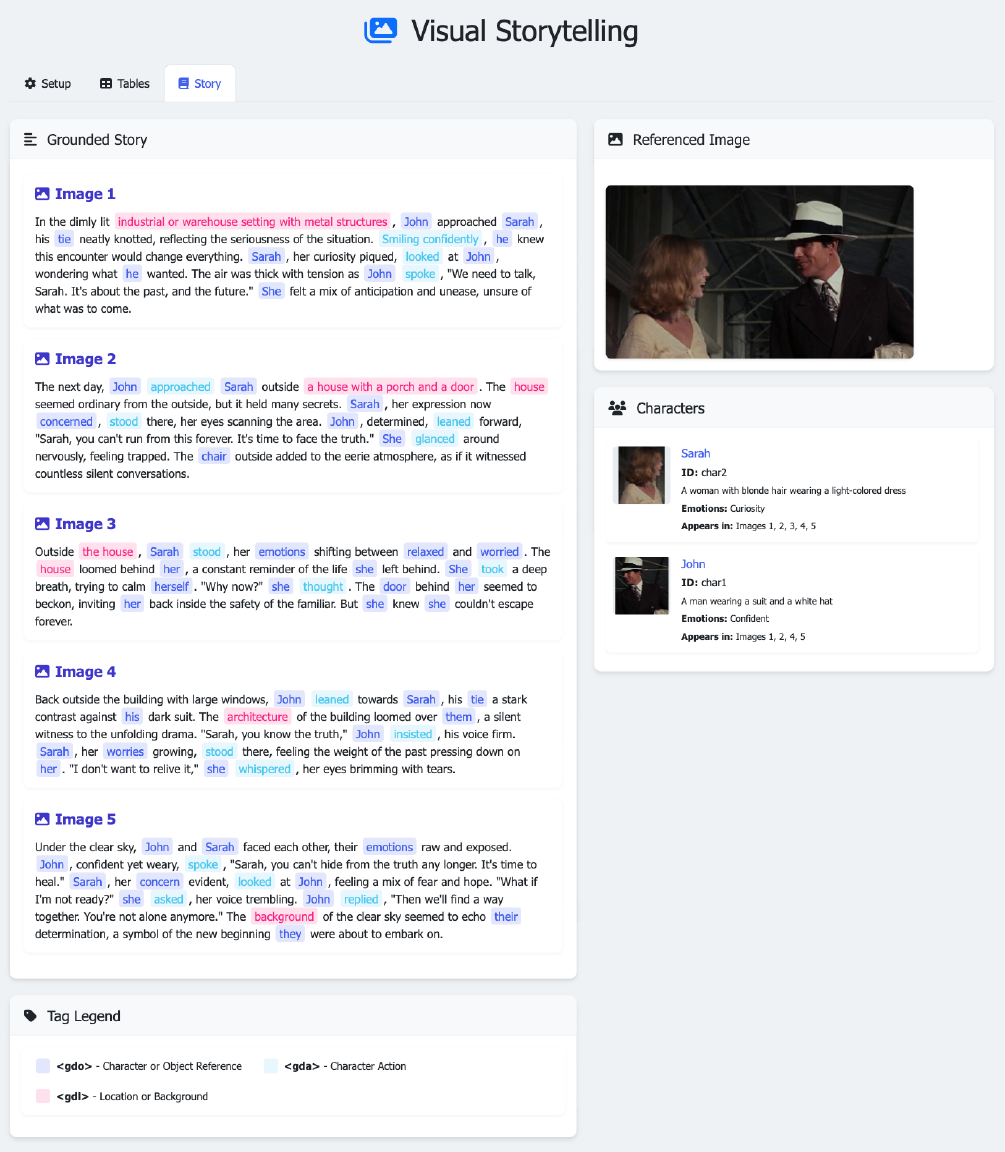}
        \caption{Interface for visualizing the grounded story in real-time. The left panel displays the narrative with color-coded entity tags: blue for character references, teal for actions, and pink for locations. The right panel shows the currently referenced image with highlighted bounding boxes corresponding to the selected entity. The user can hover over any tagged word to see the visual grounding and access detailed metadata about characters, objects, or settings.}
        \label{fig:story-interface}
    \end{figure}

    Fig.~\ref{fig:story-interface} presents the story visualization component, which displays the grounded narrative with color-coded entity tags.
    When users hover over tagged elements (characters, actions, or locations), the interface highlights the corresponding
    visual entity in the referenced image and displays relevant metadata in a tooltip.
    This interactive feature enables direct tracing between narrative elements and their visual counterparts, making the grounding
    explicit and user-friendly.

    The interface connects directly to models served via vLLM~\cite{kwon2023efficient}, allowing real-time generation of both \gls{cot} analysis
    and grounded stories from user-uploaded images.
    The generation process is streamed, providing immediate feedback as tables and story elements are produced.
    Users can observe the model's reasoning process through the structured \gls{cot} analysis before viewing the final grounded
    story with entity references.
    This visualization tool is available alongside our model code and dataset, enabling researchers and practitioners to experiment
    with visual storytelling in an accessible way.

    \newpage
    \clearpage
    \pagebreak[4]

    \section{Dataset Generation Prompts}\label{sec:dataset-generation-prompts}
    This section presents the prompts used to generate the structured scene analysis (Chain-of-Thought) and grounded stories in the StoryReasoning dataset.

    \subsection{Prompts for Chain-of-Thought generation}\label{subsec:cot-prompt}
    \begin{tcolorbox}[title=System Prompt, colback=gray!5, colframe=gray!50, coltitle=black, fonttitle=\bfseries, width=\textwidth]
        You are an expert film analyst. You will analyze a sequence of movie images, creating structured tables
        for characters, objects, and settings. I've provided object detections and landmark recognitions for each image.

        Important information about the detected objects:

        1. Each detected object has a unique object ID (like "0-person-0", "1-car-5")

        2. I will provide a mapping from these object IDs to consistent entity IDs, which come in four types:
        - Characters: char1, char2, etc. (for people or characters)
        - Objects: obj1, obj2, etc. (for things)
        - Landmarks: lm1, lm2, etc. (for recognizable places/landmarks)
        - Background: bg1, bg2, etc. (for background elements)

        3. Objects with the same entity ID represent the same element appearing across multiple images

        4. Bounding box coordinates are provided in pixel values (actual image dimensions) as x1,y1,x2,y2

        5. Landmarks provide additional context about the setting and location
        - Landmarks are formatted as "landmark-ID: landmark-name: x1,y1,x2,y2"
        - If landmarks appear in multiple images, they will have entity IDs in the mapping section

        6. Some objects may be incorrectly classified or detected:
        - You can override object classifications if they're clearly wrong (e.g., a "dog" that is actually a "cat")
        - You can ignore detections if no actual object exists at that location
        - Use your visual understanding to correct any detection errors
        - Only include objects in your tables that you can actually see in the images
        - If you are sure there is an object in the image but it is not detected, you can add it to the table, remember to include the bounding box
        - If you detect two entities that are the same but have different IDs, you must merge them into one entity and use only one of the IDs

        For each image, provide:\\
        \\
        \#\# Image X\\
        \#\#\# Characters\\
        | Character ID | Name | Description | Emotions | Actions | Narrative Function | Bounding Box\\
        IMPORTANT: If no characters are detected, you can omit the table and the section title. Bounding box is required.\\
        \#\#\# Objects\\
        | Object ID | Description | Function | Interaction | Narrative Function | Bounding Box\\
        IMPORTANT: The landmarks and background elements should be included in the objects table.
        If no objects, landmarks, or background elements are detected, you can omit the table and the section title.
        Bounding box is required.\\
        \#\#\# Setting\\
        | Setting Element | Description | Mood | Time | Narrative Function\\
        IMPORTANT: For the Setting Element column, you must use only one of these specific categories:
        Location, Environment, Lighting, Weather, Time Period, Architecture, Interior Design, Atmosphere, Background\\
        After analyzing all images, provide a Narrative Structure table connecting the images:\\
        \#\# Narrative Structure
        | Narrative Phase | Description | Key Events | Images |\\
        IMPORTANT: For the Narrative Structure table, the Narrative Phase column must use only one of these specific categories:
        Introduction, Development, Conflict, Turning Point, Conclusion
        The header \#\# Image X is required for every image even if there is no character or object table.\\
        Maintain consistent table formatting with proper markdown syntax.
        Each table must have headers and at least one row of data.
        Use the entity IDs (char1, obj2, etc.) I've provided rather than creating your own.
    \end{tcolorbox}

    \begin{tcolorbox}[title=User Message, colback=gray!5, colframe=gray!50, coltitle=black, fonttitle=\bfseries, width=\textwidth]
        \textbf{[Image Content]}\\

        Detected objects in this image:\\
        \\
        - 0-person-0: person: 125,78,347,412\\
        - 0-person-1: person: 580,102,721,389\\
        - 0-desk-0: desk: 423,310,760,415\\
        - 0-chair-0: chair: 498,190,590,309\\
        - 0-window-0: window: 42,89,120,280\\
        - 0-building-0: building: 5,5,830,450\\
        - landmark-0: Empire State Building (visible through window): 58,110,98,260\\

        Object ID mapping for this image (object ID → entity ID):\\
        \\
        - 0-person-0 → char1\\
        - 0-person-1 → char2\\
        - 0-desk-0 → obj1\\
        - 0-chair-0 → obj2\\
        - 0-window-0 → bg1\\
        - 0-building-0 → bg2\\
        - landmark-0 → lm1\\
        \\
        This is image 1 of 5.\\
        Please update your analysis with this image.
    \end{tcolorbox}

    \subsection{Prompts for Grounded Story Generation}\label{subsec:story-prompt}
    \begin{tcolorbox}[title=System Prompt, colback=gray!5, colframe=gray!50, coltitle=black, fonttitle=\bfseries, width=\textwidth]
        You are a creative storyteller who creates vivid, creative, and inspiring stories based on visual scenes.
        Using the image-by-image analysis provided, craft a compelling, creative story that's plausible given
        the visual elements, characters, and settings shown in the images.\\
        \\
        Important:\\
        - Your story should NOT be a simple description of what's in each image\\
        - Create an engaging narrative with plot, character development, and emotional depth\\
        - Feel free to invent character names, backstories, motivations, and relationships\\
        - You can create any type of story (drama, mystery, romance, adventure, etc.) that's plausible given the visual elements but it should engage the reader\\
        - Your narrative can be completely different from any original movie these images might be from\\
        - Avoid repetitions of nouns, preferring to use <gdo char1 char2 char3>They</gdo> instead of <gdo char1>John</gdo> and <gdo char2>Mary</gdo> and <gdo char3>Bob</gdo> in the same sentence\\
        - You don't have to use every object in the tables in your story, but you should use the most relevant ones\\
        \\
        Your story should use special grounding tags to reference elements from the analysis:
        \\
        1. Image grounding: <gdi image1>Text describing events in image 1</gdi>
        Each part of the story must be inside a image tag indicating which image it describes\\
        2. Character and action tags:
        For character references: <gdo char1>Character name, pronoun or description</gdo> or <gdo char1 char2>They</gdo> for multiple characters
        For character actions: <gda char1>action description</gda> or <gda char1 char2>action description</gda> for multiple characters\\
        3. Object grounding: <gdo obj1>Object reference</gdo> or <gdo obj1 obj2>Objects reference</gdo>
        Use this for specific objects in the scene\\
        4. Landmark grounding: <gdl lm1>Landmark description</gdl> or <gdl lm1 lm2>Landmarks description</gdl>
        Use this for landmarks or recognizable locations\\
        5. Background grounding: <gdl bg1>Background element description</gdl> or <gdl bg1 bg2>Background elements description</gdl>
        Use this for background elements or general settings\\
        \\
        Example of properly formatted text:\\
        <gdi image1>\\
        <gdo char1>Sarah</gdo> <gda char1>held</gda>\\
        <gdo obj3>the ancient book</gdo> as <gdo char1>she</gdo> <gda char1>gazed</gda> across\\
        <gdl lm1>the famous cliffs</gdl> rising above\\
        <gdl bg1>the misty shoreline</gdl>.\\
        </gdi>\\

        <gdi image2>\\
        The wind picked up as <gdo char1 char2>they</gdo> <gda char1 char2>walked</gda> toward\\
        <gdl lm2>the abandoned lighthouse</gdl> <gda char1 char2>standing</gda> on\\
        <gdl bg2 bg3>the rocky shoreline</gdl>.\\
        </gdi>\\

        Create a rich, inspiring or suspense story that's plausible given the visual elements in the images,
        making sure every part of the text is within the appropriate image tag.
    \end{tcolorbox}

    \begin{tcolorbox}[title=User Prompt, colback=gray!5, colframe=gray!50, coltitle=black, fonttitle=\bfseries, width=\textwidth]
        Here is the image-by-image analysis:

        [CoT analysis generated in the previous step]

        Use the following elements in your story with their corresponding tags:

        CHARACTERS:\\
        char1: person\\
        char2: person\\
        \\
        OBJECTS:\\
        obj1: desk\\
        obj2: chair\\
        \\
        LANDMARKS:\\
        lm1: Empire State Building\\
        \\
        BACKGROUND ELEMENTS:\\
        bg1: window\\
        bg2: building\\
        \\
        I'll also include the actual images for reference. Create a creative and engaging grounded story based on this analysis.
        [Sequence of Images]

        Now, based on all the images and the analysis, create the creative grounded story as described.
    \end{tcolorbox}

    \newpage
    \clearpage
    \pagebreak[4]

    \section{Example Sample}\label{sec:example-sample}
\begin{figure}[H]
        \centering
        \begin{subfigure}[t]{0.48\textwidth}
            \centering
            \includegraphics[width=\textwidth]{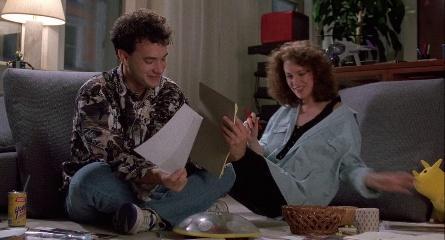}
            \caption{Image 1}
            \label{fig:sample-image1}
        \end{subfigure}
        \hfill
        \begin{subfigure}[t]{0.48\textwidth}
            \centering
            \includegraphics[width=\textwidth]{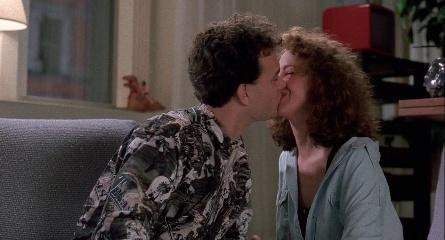}
            \caption{Image 2}
            \label{fig:sample-image2}
        \end{subfigure}

        \vspace{0.5cm}

        \begin{subfigure}[t]{0.48\textwidth}
            \centering
            \includegraphics[width=\textwidth]{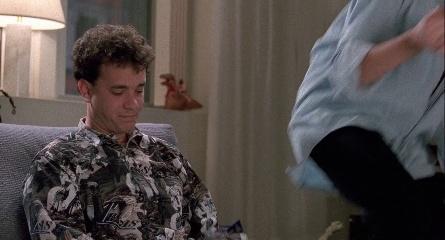}
            \caption{Image 3}
            \label{fig:sample-image3}
        \end{subfigure}
        \hfill
        \begin{subfigure}[t]{0.48\textwidth}
            \centering
            \includegraphics[width=\textwidth]{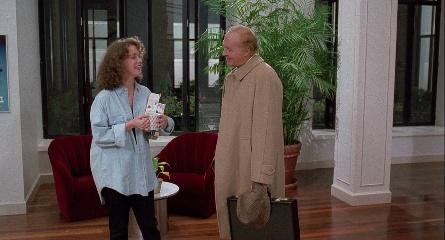}
            \caption{Image 4}
            \label{fig:sample-image4}
        \end{subfigure}

        \vspace{0.5cm}

        \begin{subfigure}[t]{0.48\textwidth}
            \centering
            \includegraphics[width=\textwidth]{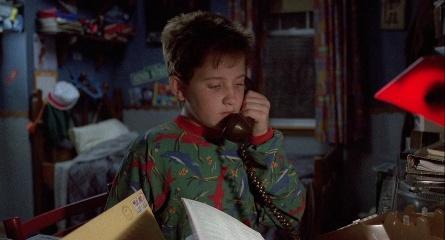}
            \caption{Image 5}
            \label{fig:sample-image5}
        \end{subfigure}

        \caption{The five-image sequence from sample 974 depicting a narrative arc with multiple characters and settings.}
        \label{fig:sample-images}
    \end{figure}

    This section shows an example sample from the StoryReasoning dataset.
    We collected all the sample with 5 images (for simplicity and space constraints) and
    randomly selected one to avoid bias towards selecting a good sample.
    The markdown tables were converted into normal tables for easier reading.
    Fig.~\ref{fig:sample-images} shows the images of this sample.
    Section~\ref{subsec:sample-chain-of-thought-(wrapped-by-<think></think>-tags)} presents the \gls{cot}
    and Section~\ref{subsec:grounded-story} presents the grounded story.
    In this specific sample we can see that in Fig.~\ref{fig:sample-image3} the model
    created a new char (char-4) likely because the person is only partially visible.

    \subsection{Chain-of-Thought (wrapped by <think></think> tags)}\label{subsec:sample-chain-of-thought-(wrapped-by-<think></think>-tags)}
    The following tables show the content from the sample converted from markdown to normal tables for easier reading.
    Bounding box columns are hidden for space constraints.

        {\large Image 1}

    \begin{table}[H]
        \centering
        \renewcommand{\arraystretch}{1.2}
        \resizebox{\textwidth}{!}{%
            \begin{tabular}{|p{0.06\textwidth}|p{0.08\textwidth}|p{0.28\textwidth}|p{0.09\textwidth}|p{0.24\textwidth}|p{0.15\textwidth}|}
                \hline
                \rowcolor[gray]{0.9} \textbf{ID} & \textbf{Name} & \textbf{Description}                                                                  & \textbf{Emotions} & \textbf{Actions} & \textbf{Function} \\
                \hline
                char1                            & Tom           & A man sitting on a sofa wearing a patterned shirt and jeans, holding a piece of paper & Engaged & Reading a paper and interacting with the woman & Protagonist \\
                \hline
                \rowcolor[gray]{0.97} char2      & Linda         & A woman sitting on a sofa wearing a green jacket, holding a piece of paper & Curious & Asking a question and pointing at the paper & Supporting character \\
                \hline
            \end{tabular}%
        }
        \caption{Characters}
        \label{tab:characters-image1}
    \end{table}

    \begin{table}[H]
        \centering
        \renewcommand{\arraystretch}{1.2}
        \resizebox{\textwidth}{!}{%
            \begin{tabular}{|p{0.06\textwidth}|p{0.15\textwidth}|p{0.15\textwidth}|p{0.20\textwidth}|p{0.34\textwidth}|}
                \hline
                \rowcolor[gray]{0.9} \textbf{ID} & \textbf{Description} & \textbf{Function} & \textbf{Interaction} & \textbf{Narrative Function} \\
                \hline
                obj1                             & Potted plant         & Decoration        & None                         & Adds to the homey atmosphere   \\
                \hline
                \rowcolor[gray]{0.97} obj2       & Curtain              & Decoration        & None                         & Adds to the cozy environment   \\
                \hline
                obj3                             & Sofa                 & Seating           & Characters are sitting on it & Provides a comfortable setting \\
                \hline
                \rowcolor[gray]{0.97} obj6       & Wicker basket        & Storage           & None                         & Adds to the homey atmosphere   \\
                \hline
                obj7                             & Mug                  & Drinkware         & None                         & Adds to the casual setting     \\
                \hline
            \end{tabular}%
        }
        \caption{Objects}
        \label{tab:objects-image1}
    \end{table}

    \begin{table}[H]
        \centering
        \renewcommand{\arraystretch}{1.2}
        \resizebox{\textwidth}{!}{%
            \begin{tabular}{|p{0.15\textwidth}|p{0.15\textwidth}|p{0.12\textwidth}|p{0.12\textwidth}|p{0.36\textwidth}|}
                \hline
                \rowcolor[gray]{0.9} \textbf{Setting Element} & \textbf{Description} & \textbf{Mood} & \textbf{Time} & \textbf{Narrative Function} \\
                \hline
                Location                                      & Living room          & Cozy          & Evening       & Establishes the setting for the interaction \\
                \hline
                \rowcolor[gray]{0.97} Lighting                & Warm indoor lighting & Comfortable   & Evening       & Enhances the intimate setting               \\
                \hline
                Interior Design                               & Casual home decor    & Inviting      & Evening       & Reflects the characters' personalities      \\
                \hline
            \end{tabular}%
        }
        \caption{Setting}
        \label{tab:setting-image1}
    \end{table}

        {\large Image 2}

    \begin{table}[H]
        \centering
        \renewcommand{\arraystretch}{1.2}
        \resizebox{\textwidth}{!}{%
            \begin{tabular}{|p{0.06\textwidth}|p{0.09\textwidth}|p{0.29\textwidth}|p{0.12\textwidth}|p{0.18\textwidth}|p{0.16\textwidth}|}
                \hline
                \rowcolor[gray]{0.9} \textbf{ID} & \textbf{Name} & \textbf{Description}                                               & \textbf{Emotions} & \textbf{Actions} & \textbf{Function} \\
                \hline
                char1                            & Tom           & A man sitting on a sofa wearing a patterned shirt, leaning forward & Affectionate & Kissing the woman & Protagonist \\
                \hline
                \rowcolor[gray]{0.97} char2      & Linda         & A woman sitting on a sofa wearing a green jacket, leaning forward & Affectionate & Kissing the man & Supporting character \\
                \hline
            \end{tabular}%
        }
        \caption{Characters}
        \label{tab:characters-image2}
    \end{table}

    \begin{table}[H]
        \centering
        \renewcommand{\arraystretch}{1.2}
        \resizebox{\textwidth}{!}{%
            \begin{tabular}{|p{0.06\textwidth}|p{0.14\textwidth}|p{0.14\textwidth}|p{0.17\textwidth}|p{0.39\textwidth}|}
                \hline
                \rowcolor[gray]{0.9} \textbf{ID} & \textbf{Description} & \textbf{Function} & \textbf{Interaction} & \textbf{Narrative Function} \\
                \hline
                obj8                             & Vase                 & Decoration        & None                 & Adds to the setting's aesthetic \\
                \hline
            \end{tabular}%
        }
        \caption{Objects}
        \label{tab:objects-image2}
    \end{table}

    \begin{table}[H]
        \centering
        \renewcommand{\arraystretch}{1.2}
        \resizebox{\textwidth}{!}{%
            \begin{tabular}{|p{0.15\textwidth}|p{0.15\textwidth}|p{0.12\textwidth}|p{0.12\textwidth}|p{0.36\textwidth}|}
                \hline
                \rowcolor[gray]{0.9} \textbf{Setting Element} & \textbf{Description} & \textbf{Mood} & \textbf{Time} & \textbf{Narrative Function} \\
                \hline
                Location                                      & Living room          & Intimate      & Evening       & Establishes the setting for the romantic moment \\
                \hline
                \rowcolor[gray]{0.97} Lighting                & Warm indoor lighting & Romantic      & Evening       & Enhances the intimate atmosphere                \\
                \hline
            \end{tabular}%
        }
        \caption{Setting}
        \label{tab:setting-image2}
    \end{table}

        {\large Image 3}

    \begin{table}[H]
        \centering
        \renewcommand{\arraystretch}{1.2}
        \resizebox{\textwidth}{!}{%
            \begin{tabular}{|p{0.06\textwidth}|p{0.09\textwidth}|p{0.29\textwidth}|p{0.10\textwidth}|p{0.24\textwidth}|p{0.12\textwidth}|}
                \hline
                \rowcolor[gray]{0.9} \textbf{ID} & \textbf{Name} & \textbf{Description}                                                 & \textbf{Emotions} & \textbf{Actions} & \textbf{Function} \\
                \hline
                char1                            & Tom           & A man sitting on a sofa wearing a patterned shirt, looking surprised & Surprised & Observing the woman's action & Protagonist \\
                \hline
                \rowcolor[gray]{0.97} char4      & Woman         & A woman standing up wearing black pants and a blue top & Angry & Standing up abruptly & Antagonist \\
                \hline
            \end{tabular}%
        }
        \caption{Characters}
        \label{tab:characters-image3}
    \end{table}

    \begin{table}[H]
        \centering
        \renewcommand{\arraystretch}{1.2}
        \resizebox{\textwidth}{!}{%
            \begin{tabular}{|p{0.06\textwidth}|p{0.14\textwidth}|p{0.14\textwidth}|p{0.17\textwidth}|p{0.39\textwidth}|}
                \hline
                \rowcolor[gray]{0.9} \textbf{ID} & \textbf{Description} & \textbf{Function} & \textbf{Interaction} & \textbf{Narrative Function} \\
                \hline
                obj2                             & Curtain              & Decoration        & None                 & Adds to the setting's aesthetic \\
                \hline
            \end{tabular}%
        }
        \caption{Objects}
        \label{tab:objects-image3}
    \end{table}

    \begin{table}[H]
        \centering
        \renewcommand{\arraystretch}{1.2}
        \resizebox{\textwidth}{!}{%
            \begin{tabular}{|p{0.15\textwidth}|p{0.15\textwidth}|p{0.12\textwidth}|p{0.12\textwidth}|p{0.36\textwidth}|}
                \hline
                \rowcolor[gray]{0.9} \textbf{Setting Element} & \textbf{Description} & \textbf{Mood} & \textbf{Time} & \textbf{Narrative Function} \\
                \hline
                Location                                      & Living room          & Unexpected    & Evening       & Highlights the sudden change in the scene  \\
                \hline
                \rowcolor[gray]{0.97} Lighting                & Warm indoor lighting & Neutral       & Evening       & Maintains consistency with previous scenes \\
                \hline
            \end{tabular}%
        }
        \caption{Setting}
        \label{tab:setting-image3}
    \end{table}

        {\large Image 4}

    \begin{table}[H]
        \centering
        \renewcommand{\arraystretch}{1.2}
        \resizebox{\textwidth}{!}{%
            \begin{tabular}{|p{0.06\textwidth}|p{0.14\textwidth}|p{0.23\textwidth}|p{0.09\textwidth}|p{0.23\textwidth}|p{0.15\textwidth}|}
                \hline
                \rowcolor[gray]{0.9} \textbf{ID} & \textbf{Name}     & \textbf{Description}                                    & \textbf{Emotions} & \textbf{Actions} & \textbf{Function} \\
                \hline
                char2                            & Linda             & A woman wearing a light blue jacket, holding a document & Worried           & Discussing something with the man & Supporting character \\
                \hline
                \rowcolor[gray]{0.97} char5      & Man with suitcase & A man wearing a beige coat, holding a suitcase & Concerned & Discussing something with the woman & Antagonist \\
                \hline
            \end{tabular}%
        }
        \caption{Characters}
        \label{tab:characters-image4}
    \end{table}

    \begin{table}[H]
        \centering
        \renewcommand{\arraystretch}{1.2}
        \resizebox{\textwidth}{!}{%
            \begin{tabular}{|p{0.06\textwidth}|p{0.13\textwidth}|p{0.11\textwidth}|p{0.23\textwidth}|p{0.37\textwidth}|}
                \hline
                \rowcolor[gray]{0.9} \textbf{ID} & \textbf{Description} & \textbf{Function} & \textbf{Interaction} & \textbf{Narrative Function} \\
                \hline
                obj1                             & Potted plant         & Decoration        & None                   & Adds to the setting's aesthetic        \\
                \hline
                \rowcolor[gray]{0.97} obj4       & Armchair             & Seating           & Linda is leaning on it & Provides a context for the interaction \\
                \hline
                obj10                            & Suitcase             & Travel            & Held by the man        & Represents a potential departure       \\
                \hline
            \end{tabular}%
        }
        \caption{Objects}
        \label{tab:objects-image4}
    \end{table}

    \begin{table}[H]
        \centering
        \renewcommand{\arraystretch}{1.2}
        \resizebox{\textwidth}{!}{%
            \begin{tabular}{|p{0.14\textwidth}|p{0.17\textwidth}|p{0.10\textwidth}|p{0.10\textwidth}|p{0.39\textwidth}|}
                \hline
                \rowcolor[gray]{0.9} \textbf{Setting Element} & \textbf{Description}  & \textbf{Mood} & \textbf{Time} & \textbf{Narrative Function} \\
                \hline
                Location                                      & Lobby or waiting area & Neutral       & Daytime       & Suggests a change of environment and potential conflict \\
                \hline
                \rowcolor[gray]{0.97} Architecture            & Modern design         & Professional  & Daytime       & Adds to the setting's formality                         \\
                \hline
            \end{tabular}%
        }
        \caption{Setting}
        \label{tab:setting-image4}
    \end{table}

        {\large Image 5}

    \begin{table}[H]
        \centering
        \renewcommand{\arraystretch}{1.2}
        \resizebox{\textwidth}{!}{%
            \begin{tabular}{|p{0.06\textwidth}|p{0.11\textwidth}|p{0.29\textwidth}|p{0.10\textwidth}|p{0.18\textwidth}|p{0.16\textwidth}|}
                \hline
                \rowcolor[gray]{0.9} \textbf{ID} & \textbf{Name} & \textbf{Description}                                                     & \textbf{Emotions} & \textbf{Actions} & \textbf{Function} \\
                \hline
                char3                            & Young boy     & A boy wearing a patterned pajama sitting at a desk, talking on the phone & Confused & Talking on the phone & Supporting character \\
                \hline
            \end{tabular}%
        }
        \caption{Characters}
        \label{tab:characters-image5}
    \end{table}

    \begin{table}[H]
        \centering
        \renewcommand{\arraystretch}{1.2}
        \resizebox{\textwidth}{!}{%
            \begin{tabular}{|p{0.06\textwidth}|p{0.13\textwidth}|p{0.11\textwidth}|p{0.23\textwidth}|p{0.37\textwidth}|}
                \hline
                \rowcolor[gray]{0.9} \textbf{ID} & \textbf{Description} & \textbf{Function} & \textbf{Interaction} & \textbf{Narrative Function} \\
                \hline
                obj2                             & Curtain              & Decoration        & None                     & Adds to the setting's aesthetic        \\
                \hline
                \rowcolor[gray]{0.97} obj5       & Desk chair           & Seating           & The boy is sitting on it & Provides a context for the interaction \\
                \hline
                obj11                            & Cardboard box        & Storage           & On the desk              & Suggests a cluttered environment       \\
                \hline
            \end{tabular}%
        }
        \caption{Objects}
        \label{tab:objects-image5}
    \end{table}

    \begin{table}[H]
        \centering
        \renewcommand{\arraystretch}{1.2}
        \resizebox{\textwidth}{!}{%
            \begin{tabular}{|p{0.14\textwidth}|p{0.17\textwidth}|p{0.10\textwidth}|p{0.10\textwidth}|p{0.39\textwidth}|}
                \hline
                \rowcolor[gray]{0.9} \textbf{Setting Element} & \textbf{Description}       & \textbf{Mood} & \textbf{Time} & \textbf{Narrative Function} \\
                \hline
                Location                                      & Bedroom or study           & Personal      & Evening       & Provides a private setting for the boy's conversation    \\
                \hline
                \rowcolor[gray]{0.97} Interior Design         & Cluttered desk and posters & Busy          & Evening       & Reflects the boy's personality and current state of mind \\
                \hline
            \end{tabular}%
        }
        \caption{Setting}
        \label{tab:setting-image5}
    \end{table}

        {\large Narrative Structure}
    \begin{table}[H]
        \centering
        \renewcommand{\arraystretch}{1.2}
        \resizebox{\textwidth}{!}{%
            \begin{tabular}{|p{0.14\textwidth}|p{0.33\textwidth}|p{0.33\textwidth}|p{0.10\textwidth}|}
                \hline
                \rowcolor[gray]{0.9} \textbf{Narrative Phase} & \textbf{Description}                                                                                                     & \textbf{Key Events} & \textbf{Images} \\
                \hline
                Introduction                                  & Setting up the initial scene with Tom and Linda interacting in a cozy living room & Tom and Linda are reading and discussing a document & Image 1 \\
                \hline
                \rowcolor[gray]{0.97} Development             & The relationship between Tom and Linda is further explored, showing a romantic moment & Tom and Linda share a kiss & Image 2 \\
                \hline
                Conflict                                      & Tension arises as a woman (char4) appears and confronts Tom, leading to a shift in dynamics & The woman stands up abruptly, and Tom looks surprised & Image 3 \\
                \hline
                \rowcolor[gray]{0.97} Turning Point           & The setting changes to a lobby where Linda and a man with a suitcase discuss something, hinting at a potential departure & Linda talks to the man with the suitcase & Image 4 \\
                \hline
                Conclusion                                    & The narrative shifts to a young boy (char3) in his room, talking on the phone, possibly connecting the previous events & The boy is engaged in a phone call, suggesting a resolution or new development & Image 5 \\
                \hline
            \end{tabular}%
        }
        \caption{Narrative Structure}
        \label{tab:narrative-structure}
    \end{table}

    \subsection{Grounded Story}\label{subsec:sample-grounded-story}

    \texttt{<gdi image1>}\\
    In the warmth of their \texttt{<gdl bg1>}Living room\texttt{</gdl>}, \texttt{<gdo char1>}Tom\texttt{</gdo>} and \texttt{<gdo char2>}Linda\texttt{</gdo>} sat on the \texttt{<gdo obj3>}sofa\texttt{</gdo>}, surrounded by the \texttt{<gdo obj2>}curtain\texttt{</gdo>} and \texttt{<gdo obj1>}potted plant\texttt{</gdo>} that made their \texttt{<gdl bg1>}Casual home decor\texttt{</gdl>} inviting. \texttt{<gda char2>}Linda\texttt{</gda>} held a \texttt{<gdo obj6>}wicker basket\texttt{</gdo>} near her, a \texttt{<gdo obj7>}mug\texttt{</gdo>} within reach. \texttt{<gdo char1>}Tom\texttt{</gdo>}, with a sense of purpose, \texttt{<gda char1>}read aloud\texttt{</gda>} from a \texttt{<gdo obj1>}piece of paper\texttt{</gdo>} he held. The \texttt{<gdl bg1>}Warm indoor lighting\texttt{</gdl>} cast a comforting glow, making the room feel intimate. Yet, a sense of urgency underlay their interaction. \texttt{<gdo char2>}Linda\texttt{</gdo>} \texttt{<gda char2>}asked\texttt{</gda>}, "Is this really necessary, Tom?" Her voice was laced with curiosity and a hint of apprehension. \texttt{<gdo char1>}Tom\texttt{</gdo>} \texttt{<gda char1>}replied\texttt{</gda>} firmly, "We have to do this. It's the only way to secure our future." \texttt{<gdo char2>}She\texttt{</gdo>} \texttt{<gda char2>}nodded\texttt{</gda>}, unsure but trusting \texttt{<gdo char1>}him\texttt{</gdo>}.\\
    \texttt{</gdi>}

    \texttt{<gdi image2>}\\
    The room's atmosphere shifted as \texttt{<gdo char1>}Tom\texttt{</gdo>} and \texttt{<gdo char2>}Linda\texttt{</gdo>} \texttt{<gda char1 char2>}leaned forward\texttt{</gda>}, their \texttt{<gdo char1>}eyes\texttt{</gdo>} locking with each other's. The \texttt{<gdo obj8>}vase\texttt{</gdo>} in the corner seemed to witness the tender moment. \texttt{<gdo char1>}Tom\texttt{</gdo>}, driven by affection, \texttt{<gda char1>}kissed\texttt{</gda>} \texttt{<gdo char2>}Linda\texttt{</gdo>}. \texttt{<gdo char2>}She\texttt{</gdo>} \texttt{<gda char2>}responded\texttt{</gda>} with equal passion, their love serving as a beacon of hope in the face of their shared challenges. The \texttt{<gdl bg1>}Warm indoor lighting\texttt{</gdl>} enhanced the romantic ambiance. In this moment of intimacy, they found solace and strength in each other's embrace.\\
    \texttt{</gdi>}

    \texttt{<gdi image3>}\\
    The peace was abruptly shattered as a \texttt{<gdo char4>}Woman\texttt{</gdo>}, her face etched with anger, stormed into the room. \texttt{<gdo char1>}Tom\texttt{</gdo>}, caught off guard, \texttt{<gda char1>}watched\texttt{</gda>} as \texttt{<gdo char4>}she\texttt{</gdo>} \texttt{<gda char4>}stood up abruptly\texttt{</gda>}. The \texttt{<gdl bg1>}Living room\texttt{</gdl>} seemed to close in around them as the tension escalated. \texttt{<gdo char4>}She\texttt{</gdo>} \texttt{<gda char4>}shouted\texttt{</gda>}, "You can't do this! You're making a terrible mistake!" \texttt{<gdo char1>}Tom\texttt{</gdo>}, surprised and confused, \texttt{<gda char1>}tried\texttt{</gda>} to respond, but the words failed \texttt{<gdo char1>}him\texttt{</gdo>}. \texttt{<gdo char4>}She\texttt{</gdo>} \texttt{<gda char4>}glared\texttt{</gda>} at \texttt{<gdo char1>}him\texttt{</gdo>}, her presence a stark reminder of the complications in their lives.\\
    \texttt{</gdi>}

    \texttt{<gdi image4>}\\
    The scene shifted to a \texttt{<gdl bg1>}Lobby or waiting area\texttt{</gdl>} with \texttt{<gdl bg1>}Modern design\texttt{</gdl>}. Here, \texttt{<gdo char2>}Linda\texttt{</gdo>}, now in a \texttt{<gdo obj4>}light blue jacket\texttt{</gdo>}, held a \texttt{<gdo obj1>}document\texttt{</gdo>} and \texttt{<gda char2>}leaned\texttt{</gda>} against an \texttt{<gdo obj4>}armchair\texttt{</gdo>}. A \texttt{<gdo char5>}Man with a suitcase\texttt{</gdo>} stood beside \texttt{<gdo char2>}her\texttt{</gdo>}, his \texttt{<gdo obj10>}suitcase\texttt{</gdo>} a symbol of impending change. \texttt{<gdo char2>}Linda\texttt{</gdo>} \texttt{<gda char2>}spoke\texttt{</gda>}, her voice filled with worry, "I don't know if I can go through with this." \texttt{<gdo char5>}He\texttt{</gdo>} \texttt{<gda char5>}replied\texttt{</gda>}, "It's not too late to change your mind, Linda. This could be your chance to start over." The \texttt{<gdl bg1>}Architecture\texttt{</gdl>} of the lobby reflected the formal atmosphere of their conversation. \texttt{<gdo char2>}Linda\texttt{</gdo>} \texttt{<gda char2>}wondered\texttt{</gda>} if the decision \texttt{<gdo char2>}she\texttt{</gdo>} was about to make would jeopardize everything \texttt{<gdo char2>}she\texttt{</gdo>} cared for. The \texttt{<gdo obj1>}potted plant\texttt{</gdo>} nearby seemed to mirror the tension in the air.\\
    \texttt{</gdi>}

    \texttt{<gdi image5>}\\
    In the quiet of a \texttt{<gdl bg1>}Bedroom or study\texttt{</gdl>}, \texttt{<gdo char3>}Young boy\texttt{</gdo>} sat in a \texttt{<gdo obj5>}desk chair\texttt{</gdo>} and \texttt{<gda char3>}talked\texttt{</gda>} on the \texttt{<gdo obj1>}phone\texttt{</gdo>}. The \texttt{<gdl bg1>}Cluttered desk\texttt{</gdl>} and \texttt{<gdo obj11>}cardboard box\texttt{</gdo>} on the desk suggested a personal space filled with \texttt{<gdo char3>}his\texttt{</gdo>} thoughts and worries. \texttt{<gdo char3>}He\texttt{</gdo>} \texttt{<gda char3>}asked\texttt{</gda>}, "Mom, what's going on? Why is everyone acting so strange?" The \texttt{<gdl bg1>}Warm indoor lighting\texttt{</gdl>} provided a sense of security, but \texttt{<gdo char3>}he\texttt{</gdo>} \texttt{<gda char3>}felt\texttt{</gda>} the weight of uncertainty. \texttt{<gdo char3>}He\texttt{</gdo>} \texttt{<gda char3>}heard\texttt{</gda>} \texttt{<gdo char2>}Linda\texttt{</gdo>}'s voice on the other end, "Everything is going to be okay, sweetie. Just trust me." \texttt{<gdo char3>}He\texttt{</gdo>} \texttt{<gda char3>}hung up\texttt{</gda>}, unsure of what to believe. The \texttt{<gdl bg1>}Location\texttt{</gdl>} felt like a sanctuary, yet \texttt{<gdo char3>}he\texttt{</gdo>} \texttt{<gda char3>}realized\texttt{</gda>} that \texttt{<gdo char3>}his\texttt{</gdo>} life was about to change, and \texttt{<gdo char3>}he\texttt{</gdo>} \texttt{<gda char3>}didn't know how\texttt{</gda>} to prepare for it.\\
    \texttt{</gdi>}

\end{document}